\documentclass{article} % For LaTeX2e
\usepackage{iclr2025_conference,times}

\usepackage{amsmath,amsfonts,bm}

\def\eqref#1{equation~\ref{#1}}
\def\1{\bm{1}}

\DeclareMathAlphabet{\mathsfit}{\encodingdefault}{\sfdefault}{m}{sl}
\SetMathAlphabet{\mathsfit}{bold}{\encodingdefault}{\sfdefault}{bx}{n}

\usepackage{hyperref}
\usepackage{url}

\usepackage{booktabs}
\usepackage{graphicx}
\usepackage{siunitx}
\usepackage{caption}
\usepackage{float}
\usepackage{wrapfig}
\usepackage{multirow}
\usepackage{multicol}

\usepackage{xcolor}

\usepackage{subcaption} % For side-by-side figures

\usepackage{tikz}  
\usepackage{pgfplots}
\usepackage{xcolor}

\pgfplotsset{compat=1.18}  % Use the latest pgfplots version

\usepackage[nameinlink]{cleveref}

\title{Compute-Constrained Data Selection}

\author{%
  Junjie Oscar Yin\\
  Johns Hopkins University\\
  \texttt{jyin27@jhu.edu} \\
  \And
  Alexander M. Rush\\
  Cornell University\\
  \texttt{arush@cornell.edu} \\
}
\iclrfinalcopy % Uncomment for camera-ready version, but NOT for submission.
\begin{document}

\def\llama{\textsc{Llama-2 }}

\maketitle

\begin{abstract}
\label{abstract-0}

Data selection can reduce the amount of training data needed to finetune LLMs; however, the efficacy of data selection scales directly with its compute. Motivated by the practical challenge of compute-constrained finetuning, we consider the setting in which both the cost of selecting data and training are budgeted for. 
We first formalize the problem of data selection with a cost-aware utility function, 
and model the data selection problem as trading off initial-selection cost for training gain. 
We run a comprehensive sweep of experiments across multiple tasks, varying compute budget by scaling finetuning tokens, model sizes, and data selection compute. Interestingly we find that many powerful data selection methods are almost never compute-optimal, and that cheaper data selection alternatives dominate both from a theoretical and empirical perspective. For compute-optimal training, we find that perplexity and gradient data selection require training-to-selection model size ratios of 5x and 10x, respectively.

%Despite the importance of compute in data selection, little research have sought to understand performances of various data selection methods under a certain compute budget. 
%Moreover, current work largely provides negative results on methods that are relatively cheap and fast to run on. 

%We find that compute-optimal data selection varies as a function of the finetuning budget and the amount of useful data related to our target task. 

\end{abstract}

%% Introduction
\section{Introduction}
\label{Introduction}

The growth of large language models (LLMs) has motivated research into their resource profiles. The compute cost of training LLMs is substantial, and, in many cases, the total compute budget is predetermined: the number of accelerators and their usage hours are allocated in advanced. Thus, it is critical to determine the optimal allocation of resources under a budget constraint. 
Past work on compute-optimal LLMs \citep{hoffmann2022training} studies if one could attain a better perplexity for a given pretraining compute budget by balancing architecture and training decisions.

Similar resource questions exist during post-training finetuning of LLMs. We assume a common setting where a single-base LLM needs to be finetuned for a downstream task. Numerous works have sought to induce certain abilities by training from large instruction tuning datasets \citep{sanh2021multitask, wei2021finetuned, mishra2021cross}. There are several different resource constraints which might make this challenging. For example, parameter-efficient fine-tuning methods like LoRA~\citep{hu2021lora} aim to reduce memory usage during finetuning by updating only a small subset of the model's parameters. In this work we focus instead on compute-constrained finetuning. 

A promising approach to reducing compute requirements for finetuning is \textit{data selection}. Data selection is a foundational approach in machine learning where the objective is to create a minimal dataset from a collection of data~\citep{ hart1968condensed, john1975d}. Given the large computational cost of each gradient step, reducing dataset size is an appealing way to reduce this resource usage. Moreover, as larger and more diverse instruction tuning collections become available, likely only a subset of the data provides the most value for any given task. Recent work has shown that careful data selection can vastly increase the effectiveness of finetuning per step~\citep{chen2023alpagasus, zhou2024lima}. 

Yet, even if data selection is effective, it does not \textit{a priori} imply that it is compute-optimal. Given the base-level compute effectiveness of gradient descent on LLM models, data selection methods need to improve upon standard training in proportion to their added cost. 
In other words, a compute-optimal method should both improve training and be cheap to compute. In this work, we study this setting of compute-constrained data selection, and argue that this is a critical factor for practical adoption that is being under considered in method development. 

Concretely, we aim to quantify the trade off between model size, number of tokens, and data selection in LLM finetuning, such that practitioners can make well-informed decisions when choosing how to best allocate compute. We first formalize this problem as a compute-constrained combinatorial optimization problem and then discuss our compute-aware modification. We develop an categorization of different approaches for this task and model their compute scaling. This is used to frame a broad series of empirical training runs under varying compute constraints, scaling both model sizes and training length. We train over 600 models, ranging from 7 to 70 billion parameters, across 6 data selection methods and 3 downstream tasks, recording final task performances for each. 

The results from this study argue that complex data selection methods are almost never Pareto-optimal in the compute-constrained setting and that simple statistical methods such as sparse retrieval should be preferred. In particular, powerful data selection methods that use model perplexity or gradient information tend to be FLOP inefficient both from a theoretical and empirical perspective. That is not to say these methods are ineffective; for example, they should be used in settings with repeated training with different tasks on the same underlying models. We fit parameteric models to quantify the effectiveness of various approaches in a compute-aware manner. Using these fits, we extrapolate and empirically validate that, for comptue-optimal finetuning, perplexity and gradient data selection require training-to-selection model size ratios of 5x and 10x, respectively.

We hope that this framework and setting can motivate further research into cheaper data selection methods that can produce better models with less compute. \ificlrfinal Codebase and datasets to reproduce our results are available at \href{https://github.com/oseyosey/CCDS}{https://github.com/oseyosey/CCDS}. \fi
\section{Related Work}
\label{Related-Work}

% TODO: Write a introduction that summarize the contribution? Scaling laws in fine-tuning settings with data selection methods. None of these paper do the full problem, yet each tackle one facet of the problem. (then in each paragraph, shows how we go beyond)

% \colorbox{yellow}{\textbf{Replace Section Scaling laws with Compute Optimal Language Models}?}

\textbf{Compute Scaling for Model Size and Transfer Learning.} \citet{kaplan2020scaling} established the use of compute-scaling laws for language models and showed a power-law relationship between model size and loss over varying orders of magnitude. More recent works expand on the original formulations by considering learning rate schedule matching, multiple-epoch training, and hyperparameters~\citep{hoffmann2022training, muennighoff2024scaling, bi2024deepseek}. 

\citet{hernandez2021scaling} study scaling in the post-training setting but only model the relationship between pretraining and finetuning data loss. Similar work~ \citep{lin2024selecting, isik2024scaling, zhang2024scaling} studies the scaling in post-training by modeling the relation between pretraining data size, model size, finetuning method and downstream test loss. These models do not consider the use of data selection. Recent work considers the relationship between data selection and scaling in the vision domain~\citep{goyal2024scaling}, but this work does not consider the compute needed for data selection in their scaling analysis.

% Unlike our work, their scaling studies model the downstream cross-entropy loss or perplexity as target metric. 
% We scaling laws that model directly the downstream task performance of multiple target tasks such as MMLU score, where the distributions of the finetuning and test datasets need not be the same. Further, 

% Our work focuses on the trade off between pre-trained model sizes, finetuning data size, and data selection costs. 

% Our focus was on how downstream performance scales with fine-tuning compute, number of training tokens, and data selection. 

% \textbf{Our work differs from in three important ways.}
% First,

% Second,

\textbf{Data Selection for Language Models.} Data selection takes the full training data as input and chooses a subset to train~\citep{albalak2024survey}. It can be viewed as a coreset selection problem \citep{mirzasoleiman2020coresets, killamsetty2021glister}, which aims to select a subset from the given training dataset such that the model achieves performance similar to the full dataset. 

There are many different approaches to LLM data selection. The simplest are non-model specific approaches such as manual scoring functions~\citep{chen2024skill}, surface level features~\citep{robertson2009probabilistic}, and n-gram features~\citep{xie2023data}. On the other hand, more effective methods use LLMs to assign utility.  One class uses model forward inference information such as  utility scores from generations, model embeddings, and perplexity~\citep{wettig2024qurating, marion2023less, ivison2022data}. Another class uses model gradients to define influence function style selections~\citep{killamsetty2021grad, han2023context, xia2024less}. See \Cref{Data-Selection-Method-3} for further description.

% For all methods, their experiments fixed a particular data budget when comparing with baseline methods. However, this poses a problem as more advanced data selection methods use more compute budget in the data selection process that is not acocunted for. Instead of comparing them under the same data budget, our work compares them at the same compute budget, where FLOPs to do both training and data selections are controlled.

% [Auxiliary data selection is the most precised term.]

% [transition]

% \textit{we want to separate or dis-entangle between instruction tuning models selected by data selection and data selection method used for instruction tuning}

%In our settings, the goal of data selection is to determine the optimal auxiliary data that improves a model’s performance on the target distribution, and because we have access to data from the target distribution the utility function for these methods is often a form of similarity between the auxiliary and target data.

\textbf{Task-Specific Finetuning from General-Purpose Instruction Datasets.} 
While data selection can be applied in any finetuning setting, it is most impactful as a method to train a targeted model from a general-purpose dataset. In the case of LLMs, this setting is commonly training a task-specific model from an ``instruction-tuning'' dataset \citep{sanh2021multitask, wei2021finetuned, mishra2021cross}.
% Several general-purpose instruction tuning datasets exist for LLM post-training~\citep{sanh2021multitask, wei2021finetuned, mishra2021cross, chung2024scaling}. Instruction-tuned models can handle a variety of possible inputs and can be applied to downstream tasks requiring either classification or open generation \citep{wang2023codet5+, wang2023far}. 

Instruction-tuned models are effective for many downstream tasks; however they require training on a very large and expensive set of data. 
To get around this issue, models have demonstrated strong results using small subsets of instruction tuning data~\citep{chen2023alpagasus, lu2023instag, zhou2024lima}. 
As datasets grow, automated measures of quality selection has become a growing focus, particularly when many targeted models are needed. 
Therefore while instruction-tuning is not the direct focus of this work, it provide a real-world applications of compute-constrained data selection.

% Supervised finetuning language models on diverse, high-quality instruction sets, along with demonstration examples, has been shown to significantly improve zero-shot performance on unseen tasks \citep{sanh2021multitask, wei2021finetuned, mishra2021cross, chung2024scaling}. 

% Instruction tuned models train on data selected by human reviewers have been shown to yield strongest results \citep{kopf2024openassistant, muennighoff2023octopack}. 
% However, these methods are not feasible at large scales due to labour expenses. 
% As the size of dataset increases, automated measures of quality selection has become a growing focus.

% As instruction datasets become larger and more diverse, the benefits of scales appear to have diminished. In response, recent methods have focused on enhancing data efficiency by optimizing data selection utility functions that emphasize quality, diversity, and difficulty. 
% While these approaches have shown encouraging progress on selected experimental setups, it remains unclear if the benefits of data efficiency outweighs the benefits of scaling model sizes or finetuning data. In our analysis, we show that advanced data selection methods is pareto-inefficient. 

% the mini-story here: 

% \textbf{We are the first to study their scaling behaviour by modelling t}

% \textbf{And we show that data selection in instruction tuning cannot be compute agnostic. }

%% Background
% \newpage
% \input{sections/Background/2 background_4}
\section{Background}
\label{Background 5}

The goal of data selection is to choose a subset of data points from a large dataset to optimize model performance on a target task. In a learning task, we are given a large training set, \( \mathcal{D} \), a target test dataset, \( \mathcal{T} \), and a validation set, \( \mathcal{V} \). Our goal is to find the optimal subset \( \mathcal{S} \subseteq \mathcal{D} \) such that the model \( \theta = T(\mathcal{S})  \) trained on \( \mathcal{S} \) maximizes the performance on \( \mathcal{T} \) under a given data constraint:
\begin{equation}
\begin{aligned}
\mathcal{S}^{*} = \underset{\mathcal{S} \subseteq \mathcal{D}}{\arg\max} \quad P(\mathcal{T}; T(\mathcal{S})) \\
\text{subject to} \quad |\mathcal{S}| \leq K,
\label{eq:equation-1}
\end{aligned}
\end{equation}
where $P$ denotes the performance of the model on the test set and \( K \) is the max cardinality.

This problem is challenging to solve in the general case, particularly without access to the test set $\mathcal{T}$.
Approaches to the problem therefore commonly make two implicit assumptions: 
 (1) the performance function, $P(\mathcal{T}; T(\mathcal{S}))$, is monotonic and submodular, non-increasing marginal utility, in the dataset chosen \citep{kirchhoff2014submodularity}, and (2) the validation set $\mathcal{V}$ is IID with the test set  $\mathcal{T}$.
Under these assumptions, we can argue for a greedy data selection approach~\citep{kirchhoff2014submodularity}. This allows decomposing the total objective, \( P(\mathcal{T}; T({\mathcal{S}})) \), by considering the contribution of individual training points to the performance on the validation set \( P(\mathcal{V}; T(\{x\})) \) for each \( x \in \mathcal{S} \).

To estimate the marginal contribution of \( x \in \mathcal{S} \), most data selection methods use a \textit{utility function} \( v(x; \mathcal{V}) \)---as a proxy to \( P(\mathcal{V}; T(\{x\})) \)---to give the utility of each data point \( x \) based on its relevance~\citep{albalak2024survey}. 
By ranking the data points \( \mathcal{D} \) based on \( v \) and selecting those that maximize the total utility within the data budget \( K \), greedy data selection aims to obtain a high-performing subset \( \mathcal{S}^{*} \).
To summarize, we consider data selection methods that target \Cref{eq:equation-1} with a two-step greedy algorithm: score all points and then select points up to the budget $K$. 

% \textit{Monotonicity}: The loss function \( L \) is monotonic with respect to the size of \( \mathcal{S} \); that is, including more training data does not increase the loss:
%    \[
%    L(\mathcal{Z}; \theta_{\mathcal{S}_2}) \leq L(\mathcal{Z}; \theta_{\mathcal{S}_1}), \quad \text{if } \mathcal{S}_1 \subseteq \mathcal{S}_2.
%    \]

%By approximating the loss as a sum of aggregated utilities over the selected dataset \( \mathcal{S} \), we have:

% \begin{equation}
% \begin{aligned}
% L(\mathcal{Z}; \theta_{\mathcal{S}})
% &\approx \sum_{x \in \mathcal{S}} v(x) 
% &= \sum_{x \in \mathcal{S}} \sum_{z \in \mathcal{Y}} v(x, z).
% \end{aligned}
% \end{equation}

%% Taxonomy of Data Selection
\section{Compute-Constrained Data Selection}
\label{Data-Selection-Method-3}
%\( C \) denotes the computational cost (in FLOPs) of a function,
While the framework presented in \Cref{Background 5} provides a general method for data selection, we argue that it is insufficient for the practical challenge of finetuning LLMs. 
The issue is that LLM finetuning is often bottlenecked by a computational budget and not a data budget. 
There are two major computational bottlenecks in this process: (1) the cost of training the model on this data ($C_T$), and (2) the cost of computing the utility function on this data ($C_v$). The true cost of $C_v$ can reduce significantly the amount of training points we can select for given computational budget . 

Assuming we at minimum require the computation of a utility function over the dataset, we can define the \textit{compute-constrained data selection} objective as

\begin{equation}
\begin{aligned}
\mathcal{S}^{*} = \underset{\mathcal{S} \subseteq \mathcal{D}}{\arg\max} \quad P(\mathcal{V}; T(\mathcal{S})) \\
\text{subject to} \quad C_{T(\mathcal{S})}  + \sum_{x \in \mathcal{D}} C_{v(x)} \leq K.
\label{eq:equation-2}
\end{aligned}
\end{equation}

Here $K$ is now the compute, e.g. maximum number of FLOPs, allocated for data selection and training, and we assume calculation of $v$ is a fixed-cost independent of optimization. 

\subsection{Compute Cost of Data Selection Utilities}

\begin{table}[ht!]
\centering
\begin{tabular}{@{}lllc}
\toprule
\textbf{Method}         & \textbf{Utility Function}                                                                 & \textbf{Computational Cost}   & $\mathbf{C_{\text{\textbf{forward}}}(x)}$                                                                                  \\ \midrule
Lexicon-Based     & $\frac{1}{|\mathcal{V}|} \sum_{x' } \text{BM25}(x, x')$                     & $c_{\text{BM25}} ( |x| + |\mathcal{V}| |x| )$ &      $\approx$ 0                            \\
Embedding-Based         & $\frac{1}{|\mathcal{V}|} \sum_{x' } \cos(\text{Emb}(x), \text{Emb}(x'))$ & $C_{\text{embed}}(x) + C_{\text{embed}}(\mathcal{V}) $ &  $\approx \epsilon$ \\
Perplexity-Based        & $\text{PPL}_{\theta_{\mathcal{V}}}(x)$                                                   & $C_{\text{forward}}(x)$ &  $\approx$ 1                                                              \\
Gradient-Based          & $\eta_t \langle \nabla_{\theta} \ell(x; \theta^t), \nabla_{\theta} \ell(\mathcal{V}; \theta^t) \rangle$ & $3 \times C_{\text{forward}}(x) + C_{\text{grad}}(\mathcal{V})$ &  $\approx$ 3                     \\ \bottomrule
\end{tabular}
\label{tab:cost-summary}
\caption{\textbf{Utility Functions and Computational Costs for Data Selection Methods.}} %add a one liner denote what we mean by C_forward(x)
\end{table}

To make these costs more tangible, we consider four classes of data selection in this work, that represent three different levels of compute.  This section summarizes their main properties, i.e. their core utility functions and computational costs. 

\paragraph{Lexicon-Based.}

Lexicon data selection methods utilize statistical properties of text to evaluate the relevance of data points without relying on deep learning models. One of the most effective lexicon-based methods is \textit{BM25} \citep{robertson2009probabilistic, silva2024improving}, which scores data points based on the frequency of terms. The utility function \( v_{\text{BM25}}(x; \mathcal{V}) \) assigns a relevance score to each data point by averaging the BM25 scores with the validation, \( v_{\text{BM25}}(x) = \frac{1}{|\mathcal{V}|} \sum_{ x' \in \mathcal{V}} \text{BM25}(x, x') \). Since the algorithm can be run with a single-core cpu, the data selection FLOPs are almost 0.

% The computational cost of computing \( v_{\text{BM25}}(x) \) for each \( x \) is approximated by:
% \begin{align}
% C_{v_{\text{BM25}}}(x) \approx c_{\text{BM25}} \left( |x| + |\mathcal{V}| \cdot |x| \right),
% \end{align}
% where \( c_{\text{BM25}} \) is a constant representing the computational cost per token for computing term frequency operations in BM25.

% This cost grows linearly with the lengths of \( x^{(i)} \) and the aggregated lengths of the subsample \( \mathcal{Z}' \). 
% Therefore, BM25 is computationally efficient for data selection when both \( x^{(i)} \) and \( z'^{(j)} \) are not excessively long.

\paragraph{Embedding-Based.}

% Specifically, a transformer-based sentence embedding model is employed to compute embeddings of the auxiliary data points and the target data points.

These methods utilize embedding models to select data points that are most similar to the target data \citep{rubin2021learning}. The utility function \( v_{\text{retrieval}}(x; \mathcal{V}) \) assigns a score to each data point \( x \) based its cosine similarity with validation data in \( \mathcal{V} \); that is, \( v_{\text{retrieval}}(x) = \frac{1}{|\mathcal{V}|} \sum_{x' \in \mathcal{V}} \cos\left( \text{Emb}(x), \text{Emb}(x') \right) \). Assuming we are using a very small model (on the order of BERT-size) and the embedding requires only one-time transformation of data points into dense vectors, the data selection FLOPs are quite small. \citet{liu2021makes} demonstrated significant performance gains when selecting in-context examples. We apply a similar method, \textit{Embed}, to the fine-tuning setting using a small T5-based dense embedding model \citep{ni2021large}.

% The computational cost of computing \( v_{\text{retrieval}}(x) \) for each \( x \) is primarily the cost of generating embeddings:
% \begin{align}
% C_{v_{\text{retrieval}}}(x) \approx C_{\text{embed}}(x) + C_{\text{embed}}(\mathcal{V}) + C_{\text{cosine}}(x, \mathcal{V}),
% \end{align}
% where \( C_{\text{embed}}(x) \) is the cost of embedding \( x \), \( C_{\text{embed}}(\mathcal{V}) \) is the cost of embedding \( \mathcal{V} \), and \( C_{\text{cosine}}(x, \mathcal{V}) \) is the cost of computing cosine similarities between embeddings.
% Since embedding operations involve forward passes through the embedding model, the cost \( C_{\text{embed}}(x^{(i)}) \) grows linearly with the length of \( x^{(i)} \). Computing cosine similarities is relatively inexpensive compared to embedding. Therefore, the total computational cost is dominated by the embedding operations, making retrieval-based data selection computationally intensive when dealing with large datasets.
\paragraph{Perplexity-Based.}

Perplexity-based data selection utilizes language models to evaluate the utility of data points based on model loss~\citep{antonello2020selecting}. The utility function \( v_{\text{ppl}}(x; \mathcal{V}) \) assigns a score to each data point \( x \) by computing the perplexity (PPL) of \( x \) under a language model \( \theta_{\mathcal{V}} \) finetuned on \( \mathcal{V} \); that is, \( v_{\text{ppl}}(x) = \text{PPL}_{\theta_{\mathcal{V}}}(x) \). \textit{Top-PPL} and \textit{Mid-PPL} have both shown improved performance and training efficiency \citep{ankner2024perplexed, marion2023less}, where \textit{Top-PPL} ranks data points with the highest perplexity scores, and \textit{Mid-PPL} does the same for points in the middle of the score distribution.

% The computational cost of computing \( v_{\text{ppl}}(x, \mathcal{V}) \) for each \( x \) is primarily the cost of performing a forward pass through the model \( \theta_{\mathcal{V}} \):
% \begin{align}
% C_{v_{\text{ppl}}}(x) \approx C_{\text{forward}}(x).
% \end{align}

% This cost grows linearly with the length of \( x^{(i)} \) and is substantial when using large language models, making perplexity-based data selection computationally expensive for large datasets.

\paragraph{Gradient-Based.}
These methods evaluate the utility of data points based on their influence on the model's loss with respect to the target data~\citep{pruthi2020estimating}. The utility function \( v_{\text{grad}}(x; \mathcal{V}) \) quantifies this influence by computing the inner product between the gradient of the loss on \( x \) and the gradient of the loss on \( \mathcal{V} \), scaled by the learning rate \( \eta_t \); that is, \( v_{\text{grad}}(x) = \eta_t \left\langle \nabla_{\theta} \ell(x; \theta^t), \nabla_{\theta} \ell(\mathcal{V}; \theta^t) \right\rangle \), where \( \nabla_{\theta} \ell(\mathcal{V}; \theta^t) = \frac{1}{|\mathcal{V}|} \sum_{x' \in \mathcal{V}}\nabla_{\theta} \ell(x'; \theta^t) \), and \( \theta^t \) are the model parameters at time step \( t \). The computational cost of computing \( v_{\text{grad}}(x) \) for each \( x \) is approximated by:
\[
C_{v_{\text{grad}}}(x) \approx C_{\text{backward}}(x) + C_{\text{grad}}(\mathcal{V}) \approx 3 \times C_{\text{forward}}(x) + C_{\text{grad}}(\mathcal{V}),
\]
where \( C_{\text{forward}}(x)\) is the cost of a forward pass for \( x^{(i)} \) on the model \( \mathcal{M} \), and \( C_{\text{grad}}(\mathcal{V}) \) is the cost of computing \( \nabla_{\theta} \ell(\mathcal{V}; \theta^t) \), computed once for all \( x \). Computing gradients involves both forward and backward passes, totaling approximately three times the cost of a forward pass. Low-rank sgradiEnt Similarity Search (\textit{LESS}) shows superior performance gain over cheaper methods and random selection \citep{xia2024less}.

While lexicon and embedding-based methods aim to select training samples similar to validation samples, perplexity and gradient-based methods focus on optimizing their effect on model loss. Different implementation of these general categories leads to different cost. A detailed cost analysis of our selected data selection methods can be found in the \Cref{Appendix-DS-FLOPS}.

% Due to the necessity of computing gradients for each data point, gradient-based data selection is computationally intensive for large datasets.

%% Method
\section{Modeling the Compute-Performance Relationship }
\label{Compute-Aware-Data-Selection}
\begin{figure}[t]
    \centering
    \resizebox{1.0\textwidth}{!}{%
    \includegraphics{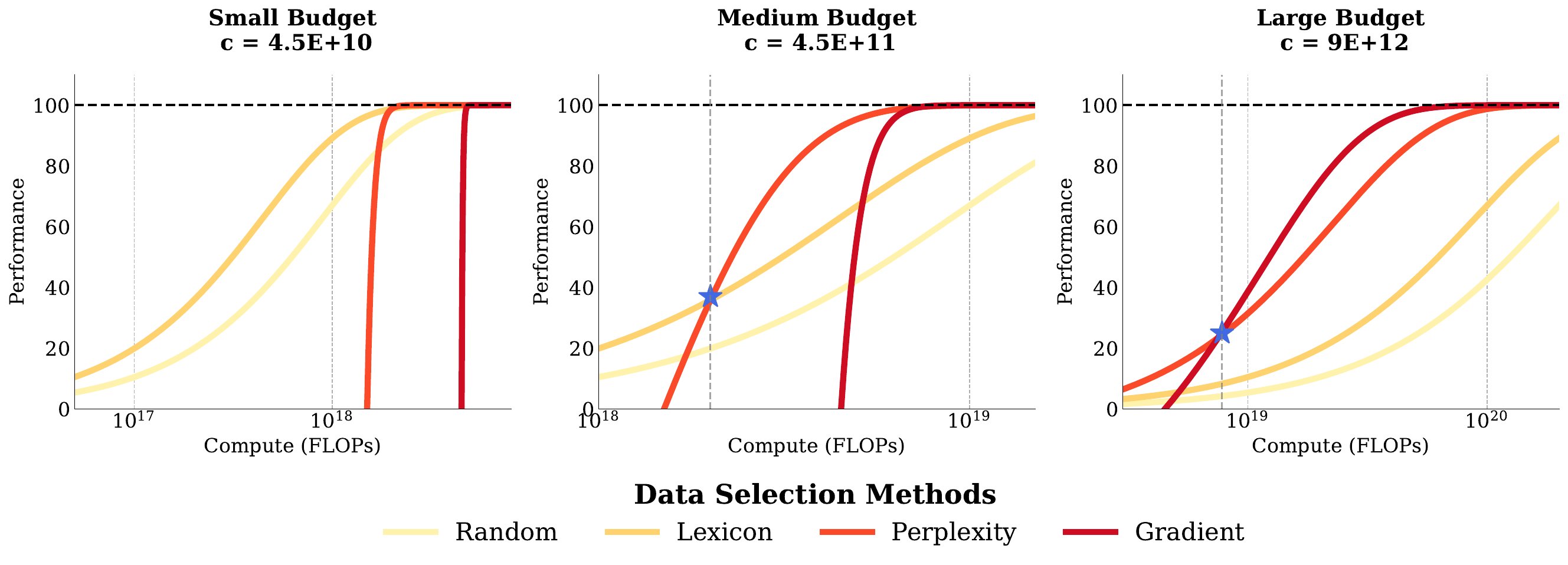} 
    }
    \caption{\textbf{Simulation of Performance under Constraints.} $P(k) = \bar{P} \times \left( 1 - \exp\left( -\lambda \frac{C(k)}{C(|\mathcal{D}|)} \right) \right)$  The behavior of different data selection methods using our performance model. 
    \textit{(Left-Small Budget)}  The Lexicon method may consistently outperform more advanced data selection methods if their initial cost is too high. Under our assumptions gradient can never be optimal as its cost exceeds 1 epoch of training.
    \textit{(Middle-Medium Budget)} The perplexity method can become optimal once the total cost exceeds a given amount.
    \textit{(Right-Large Budget)} The gradient methods can be  optimal if training is more expensive than the fixed-cost, for example if using a much larger base model than data selection model.
    The simulation shows that the compute-optimal data selection method changes as a function of the compute budget and the performance rate associated with each method.
    }
    \label{fig:simulation-gain}
\end{figure}

% The formulation in \Cref{Background 5} has a implicit goal of finding the optimal data selection cost \( C_{\text{DS}}^* \) that maximizes the expected utility \( v_{\text{DS}}(C_{\text{DS}}) \):
% \[
% C_{\text{DS}}^* = \underset{C_{\text{DS}}}{\arg\max}
% \quad \text{subject to} \quad 0 \leq C_{\text{DS}} \leq K.
% \]
 To analyze the trade-off between the compute of data selection methods and the expected gain in model performance, we define a simplified parametric form for expected peformance. Let $c$ be the fixed-cost of training on a single data point  and \( k = |\mathcal{S}| \) be the number of data points. Define \( C(k) \) as the total cost of training and selection,
\[
C(k) = c \times k + \sum_{x} C_{v(x)}.
\]

The assumption in data selection is that more compute intensive methods are able to achieve higher performance with less samples, 
but that the value of information gained from selecting additional data points diminishes as more points are explored. 
We will assume that data points are exchangeable, and that all methods eventually reach the same upper-bound performance after 1 epoch, $\bar{P} = P(\mathcal{T}; T(\mathcal{D}))$.

% Initially, selecting new data points provides significant value because each point reveals new and informative aspects of the solution space, reducing the entropy \( H(p) \). However, as more points are selected and the solution space becomes better understood, the information gained from each additional point decreases. 

Our parametric model for the expected performance after training on \( k \) data points should capture diminishing returns, dependence on computational cost of the method, and convergence toward an upper bound. We model this function as:
\begin{equation}
P(k) = (\bar{P}  - P_0)\times \left( 1 - \exp\left( -\lambda \frac{C(k)}{C(|\mathcal{D}|)} \right) \right) + P_0
\label{eq:parametric-function}
\end{equation}
where $P_0$ is the zero-shot performance,  \( \bar{P} \) is the upper bound and \( \lambda \) is the value the method extracts from additional compute in the utility function. 

\paragraph{Simulation} \Cref{fig:simulation-gain} presents a simulation of this performance function across different methods under varying compute constraints. We set $c$ to \{4.5E+10, 4.5E+11, 9E+12\}, $k$ to 100M, and utility compute to \{4.5E+10, 4.5E+11, 9E+12\}. The parameter $\lambda$ is set to \{5, 10, 40, 80\} for the Random, Lexicon, Perplexity, and Gradient methods, respectively. These settings demonstrate how different data selection methods may become compute-optimal under varying compute budgets. Note that for Medium and Large Budgets, we use a Small-size model for PPL and Gradient. 

Using methodology similar to \cite{hoffmann2022training}, the parameter \( \lambda \) can be fit on the empirical measurements, which yields data-driven estimates. See \Cref{Appendix-Parametric-Fit} for more details on derivations and the fitting procedure.

\section{Experimental Setup}

% \colorbox{yellow}{Fix Model Sizes, Data Selection Methods and Vary Number of Finetuning Tokens}

% \colorbox{pink}{CUTTING: cut into 1 paragraph.}
As shown in \Cref{tab:Experiment-Overview}, experiments vary the number of finetuning tokens for 5 data-selection methods and a fixed family of models, ranging from 7B to 70B parameters. The finetuning data budget is fixed as a percentage of the total finetuning tokens: \{2.5, 5, 10, 25, 50, 100\}\%, across 3 target tasks. For each finetuning budget, we conduct multiple training runs with increasing compute, which is either allocated toward larger pre-trained model sizes or more sophisticated data selection methods. 
% The training runs allows us to interpolate the downstream performance curves for one-epoch training. 
We analyze each FLOP count to identify which runs achieve the highest performance on target benchmark. We then fit a power law to obtain a finetuned Pareto frontier for each model sizes.
% From this, we obtain a continuous mapping from FLOP count to performance for each combination of data-selection method and model size. 

% To determine the optimal allocation of compute, we analyze each FLOP count to identify which runs achieve the highest performance on target benchmarks. Using this analysis, we derive a mapping from any given FLOP count $C$ to the most effective choice of model size $N$, number of training tokens $D$, and data-selection method $S$ such that the total computational cost, $\text{FLOPs}(N, D, S)$, equals $C$. Next, for a set of logarithmically spaced FLOP intervals $L$, we identify the model size $N$ that yields the highest performance, along with the corresponding $D$ and $S$. We then fit scaling laws to estimate the optimal model size, number of training tokens, and data-selection strategy for any given compute budget.

% Next, for a set of logarithmically spaced FLOP intervals $L$, we identify the model size $N$ that yields the highest performance, along with the corresponding $D$ and $S$. Finally, we fit power laws to estimate the optimal model size, number of training tokens, and data-selection strategy for any given compute budget.

\begin{table}[t]
\centering
\begin{tabular}{ccccc}
\toprule
\textbf{Training Data} & \textbf{Data Selection Method} & \textbf{Model Size} & \textbf{Target Task} \\
\cmidrule(lr){1-1} \cmidrule(lr){2-2} \cmidrule(lr){3-3} \cmidrule(lr){4-4}
2.5\%  & Random  & \textsc{Llama}2 7B  & MMLU  \\
5\%    & BM25    & \textsc{Llama}3 8B  & BBH   \\
10\%   & Embed   & \textsc{Llama}2 13B  & IFEval\\
25\%   & PPL     & \textsc{Llama}2 70B &       \\
50\%  & LESS    &     &       \\
100\%  &     &     &       \\
\bottomrule
\end{tabular}
\caption{\textbf{Experimental Setup Overview}.}
\label{tab:Experiment-Overview}
\end{table}

%\section{Experimental Setup}

\paragraph{Datasets}

We follow \citet{wang2023far} and curate a representative sample of instruction-tuned datasets as listed in \Cref{tab:dataset_details}. This includes: (1) datasets generated by researchers from existing NLP datasets, such as COT \citep{wei2022chain} and Flan V2 \citep{longpre2023flan} ; (2) datasets written by humans from scratch specifically for instruction tuning, including Dolly \citep{conover2023free} and Open Assistant 1 \citep{kopf2024openassistant}.

For evaluating the model we run on three challenging but different downstream tasks. These include:  the Massive Multitask Language Understanding (\textbf{MMLU}, \cite{hendrycks2020measuring}) dataset measures models' factual knowledge, comprising questions across 57 subjects, spanning difficulty levels from elementary to professional;  Big-Bench-Hard (\textbf{BBH},  \cite{suzgun2022challenging}) curates 23 complex reasoning tasks from Big-Bench \citet{srivastava2022beyond}. It is used to evaluate models' general reasoning capabilities; and Instruction Following Evaluation (\textbf{IFEval}, \cite{zhou2023instruction}), which evaluates models' instruction following abilities. For MMLU, we report 5-shot accuracy; for BBH, we report 3-shot exact match score; and for IFEval, we report 0-shot accuracy.

% We report the 3-shot exact match score across all tasks. 

% Its multiple-choice format allows for effective probing of models’ knowledge, minimizing concerns about open-ended responses. We report the 5-shot performance on MMLU. 

% \textbf{BBH}. Big-Bench-Hard (BBH \cite{suzgun2022challenging}) curates 23 complex reasoning tasks from Big-Bench \citet{srivastava2022beyond}. It is used to evaluate models' general reasoning capabilities. We report the 3-shot exact match score across all tasks. 

% \textbf{IFEval}. Instruction Following Evaluation (BBH \cite{suzgun2022challenging}) curates 23 complex reasoning tasks from Big-Bench \citet{srivastava2022beyond}. It is used to evaluate models' general reasoning capabilities. We report the 3-shot exact match score across all tasks. 

\paragraph{Pretrained Models}

For all experiments, we train transformer language models with \textsc{Llama} architecture and tokenizer \citep{touvron2023llama, dubey2024llama}. Models range from 7B parameters to 70B parameters, and are trained for close to 100 million total fine-tuning tokens. Our experiments closely follow prior work on training and evaluating instruction-tuned models \citep{wang2023far, ivison2023camels}.

We primarily focus on the \llama model suite, containing \textsc{Llama-2-7B}, \textsc{Llama-2-13B}, and \textsc{Llama-2-70B}. Because the latest \textsc{Llama}-3 model suit only contains two model sizes (8B and 70B), we believe the three \textsc{Llama}-2 models are better suited in modelling the scaling behaviour. Nevertheless, We show that our results generalizes well to \textsc{Llama}-3 by experimenting with the smaller base model \textsc{Llama-3-8B}. 

% Since we are interested in modelling the scaling behaviour of model sizes, we didn't experimented with latest \textsc{Llama}-3 model suite that only contains two model sizes (8B and 70B). On the whole, the three \llama models represent the largest, most-used, and highest-quality pretrained model best suit for a scaling study.

% Documents with a low perplexity are assumed to be close to target domain. 

% TODO: Find the appropriate citations for top-k ppl.
%Instances with low perplexity are assumed to be close to the reference domain while data with paragraphs with higher perplexity are assumed to be of low utility. 

% \input{tables/Data_Select_Methods}

% \textbf{GSM}. The Grade School Math (GSM \cite{cobbe2021training}) dataset evaluates models' mathematical reasoning abilities. GSM contains 8.5K grade school math word problems with natural language solutions. We report the 8-shot exact match score across all tasks.

% A full breakdown of results for each target task is shown in appendix.
% \input{tables/Target_task}

\begin{figure}[!t]
    \centering
    \resizebox{1.0\textwidth}{!}{%
    \includegraphics{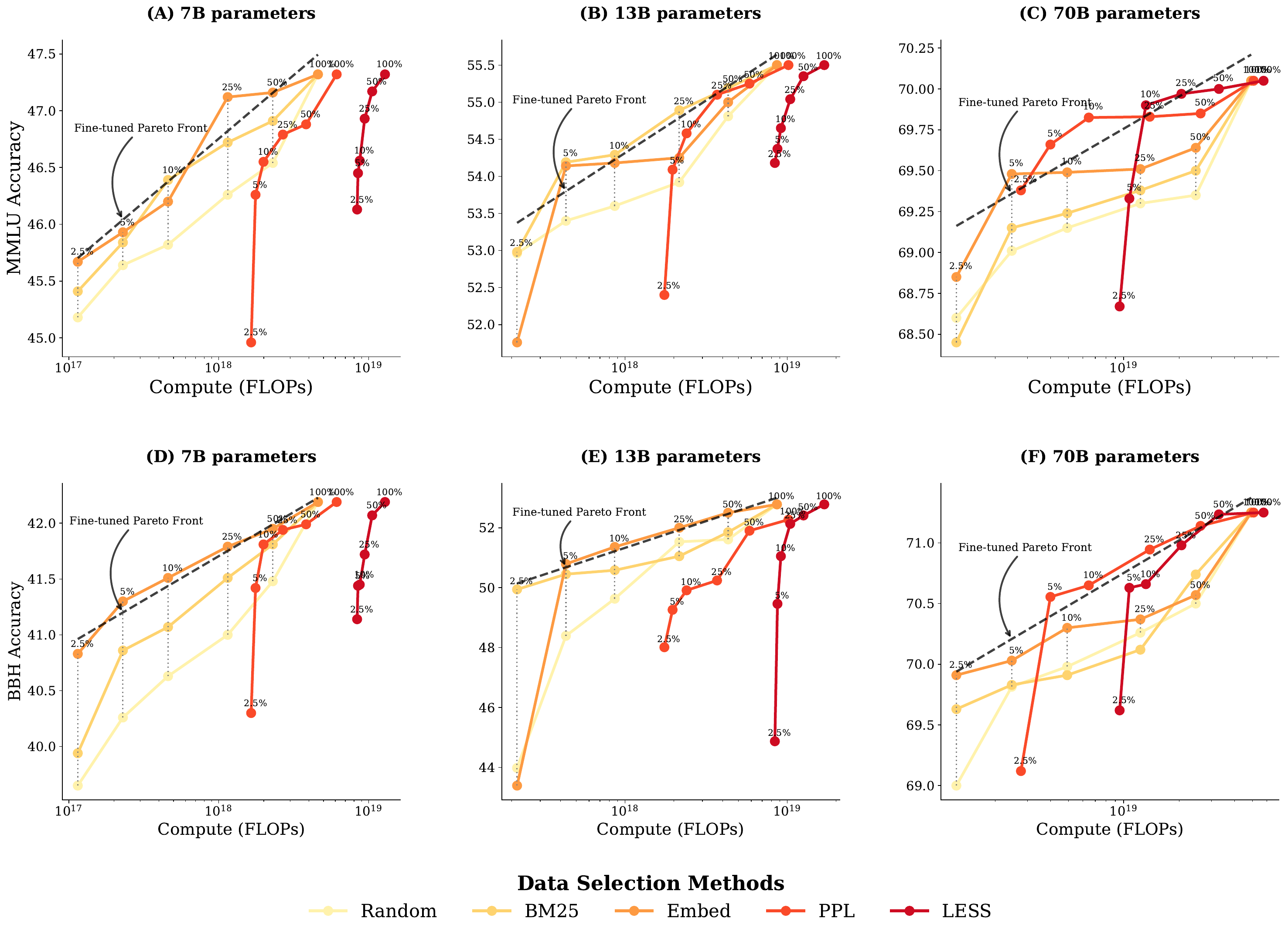} 
    }
    \caption{\textbf{Performance for Different Data Selection Methods}. We show all of our different runs for a given model size, where each scatter point is the final target task performance of a single run. \textit{(A, B, C)} show MMLU results across three model sizes, while \textit{(D, E, F)} present BBH results across three model sizes. For each run, we determine the optimal finetuning strategy---a combination of data selection method and number of finetuning tokens---that achieves the highest performance under a particular FLOPs budget. We fit a pareto front in dashed line based on these optimal strategies, which is a line in the linear-log space. At small and medium compute budgets \textit{(A, B, D, E)}, cheaper data selection methods like BM25 and EMBED outperform PPL and LESS, which rely on model information. At larger compute budgets \textit{(C, F)}, however, PPL and LESS become compute-optimal after using 5\% of the fine-tuning tokens. }
    \label{fig:allocation-experiment}
\end{figure}

\section{Results}

\paragraph{Empirical Results.}

\Cref{fig:allocation-experiment} shows the full results with 5 data selection methods across 3 pretrained model size and 2 of the target task. For these experiments, the unique training data budget is fixed at roughly \{2.5\%, 5\%, 10\%, 25\%, 50\%, 100\%\} of tokens. For each data budget, we finetune a set of models with increasing amount of compute that is allocated to either more parameters or more expensive data selection methods. 

Note that for PPL and Gradient, a 7B model of the same model family is always used for data selection; whereas for Embed a small encoder model is used. For MMLU, PPL is implemented as \textit{Mid-PPL}. For BBH, PPL is implemented as \textit{Top-PPL}.

\textit{The fine-tuned efficient Pareto frontier} comprises all runs that are Pareto-optimal with respect to compute (x\text{-axis}) and performance (y\text{-axis}). These runs represent the most efficient choices, providing the best possible performance for a given compute budget under specific data selection methods and training token lengths. Furthermore, we model the efficient computational frontier using a power law, specifically of the form $P(C) = a \log(C) + b,$ where \( a \) and \( b \) are parameters fitted to the data. The fitted function is depicted as a dashed gray line in the figures.

The main 7B results in \Cref{fig:allocation-experiment} \textit{(A, D)} and \Cref{fig:results-analysis-combined} (b), show that cheap lexicon-based methods (BM25) and embedding-based methods (Embed) significantly outperform perplexity-based (PPL) and gradient-based methods (LESS). While PPL and LESS achieve better performance at the same \textit{data budget} compared to these methods (see \Cref{fig:results-analysis-combined}), they are not compute optimal under the same \textit{compute budget} due to the high FLOPs required for data selection. The marginal benefit one can get from using a more sophisticated data selection methods does not outweigh its cost in selecting these data.  Additional results can be found in \Cref{Appendix-Additional-Results-IFEval}. 
% \colorbox{yellow}{7B IFEval analysis? One line is enough? Results similar}
% In contrast, retri methods, in particular BM25, are more compute optimal, offering the highest 'FLOP-to-performance' value among the data selection methods.

 As models scale to 13B \Cref{fig:allocation-experiment} \textit{(B, E)},
 we still find that expensive data selection methods underperform, even though, the relative cost of doing data selection for these methods diminishes as the training model size increases. As shown in \Cref{fig:allocation-experiment} \textit{(B, E)}, at 13B model sizes, cheaper methods are still preferred. We do see that after 5\% finetuning tokens, PPL is more competitive, outperforming Embed in MMLU, and almost matching the pareto front at 25\% finetuning tokens. 

At the 70B model size, shown in \Cref{fig:allocation-experiment} \textit{(C, F)}, PPL and LESS outperform both BM25 and Embed finetuning tokens for the first time. This suggests that at very large compute budgets, more sophisticated and costly methods can gain a greater advantage compared to lexical and embedding methods. As model sizes continue to scale and LLM sizes shrink, these methods may become more efficient. 

% One liner 
Results on \textsc{Llama}3 8B are nearly identical to the 7B we see for \textsc{Llama}2. This verify that the approach is not model specific. We include full results in \Cref{appendix-results-llama3}.

% One liner for IFEval

%  Overall, these results suggest that at small to medium compute budgets, the optimal \textit{allocation} strategy is to use cost-effective methods such as BM25 and EMBED. More expensive methods are not justified at these scales. However, at larger compute budgets, the relative cost of these methods diminishes, and the benefit of extracting more signal from the data outweighs the costs, rendering these methods compute-optimal.

% Overall, these results suggest that the optimal \textit{Allocation} strategy is to scale pre-trained model size first with a cost-effective data selection method. For a given model size, opting BM25 for data selection is the most efficient method. 

% Across model sizes and target tasks, our results align with our theoretical analysis that the best choice of data selection method can vary as a function of our compute budget $(K)$. 

% \colorbox{pink}{A similar analysis is expected in BBH. But it seems redundant to repeat the same analysis?}

% \section{Results: Finetuning Allocation for IFEval}

\begin{figure}[!t]
    \centering
    \resizebox{1.0\textwidth}{!}{%
    \includegraphics{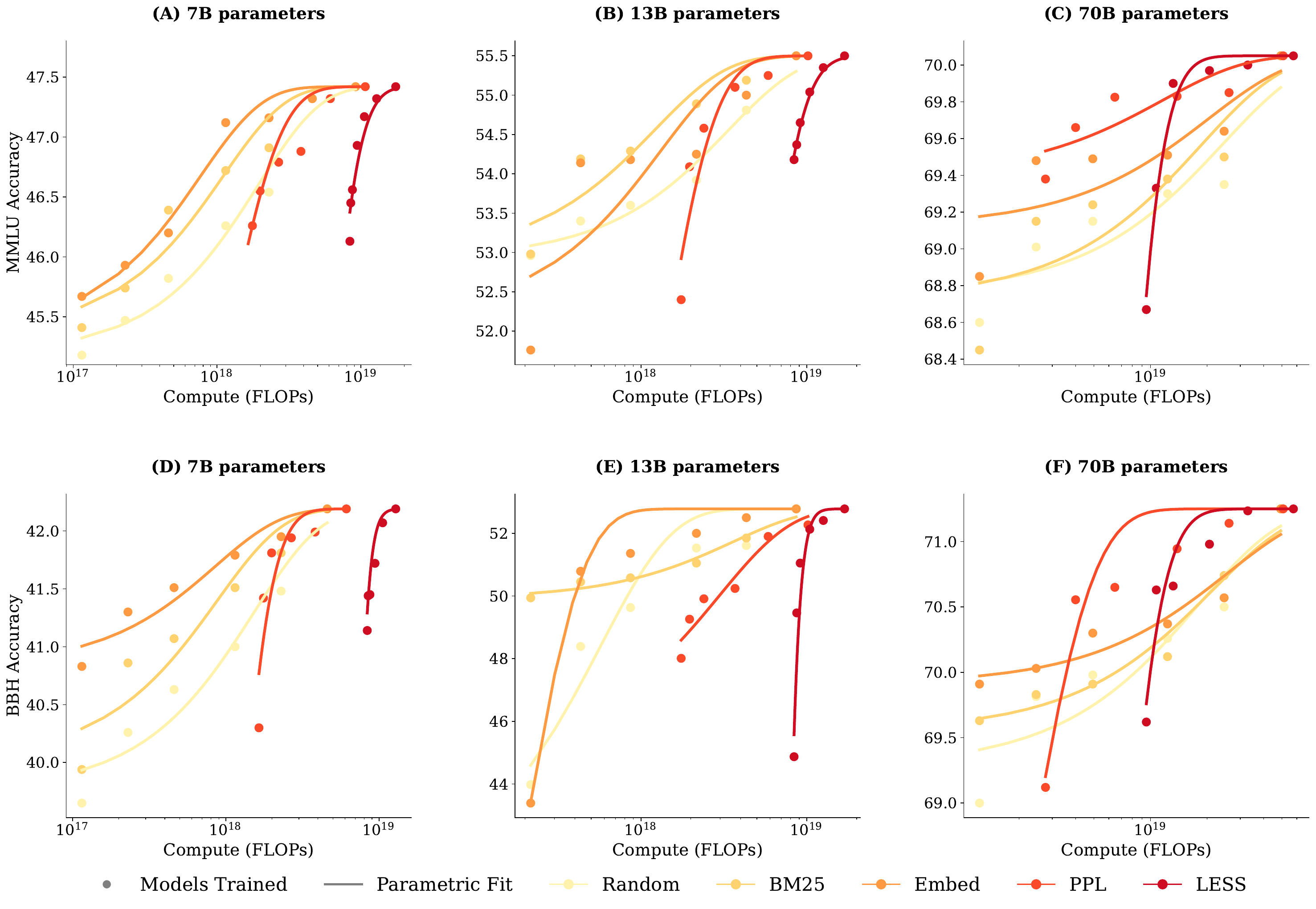} 
    }
    \caption{\textbf{Parametric Fit of Performance with Compute-Constrained Data Selection.} We fit a parametric model of the performance in \Cref{eq:parametric-function} and display that as curves to pair with the empirical results as scatter points. \textit{(A, B, C)} show MMLU results and their parametric fit across three model sizes, while \textit{(D, E, F)} present BBH results and their parametric fit across three model sizes. }
    \label{fig:parametric-fit}
\end{figure}

\textbf{Fit of Compute-Performance Relationship.}  In \Cref{Compute-Aware-Data-Selection} we propose a parameteric model for the relationship of data selection compute to model performance, \Cref{eq:parametric-function}. This formula assumed that data selection diminish in utility and that performance is a direct function of compute. 
To verify these assumption, we fit the parametic form to the the empirical curves by fitting the $\lambda$ parameter and allowing tolerance in $P_0, \bar{P}$, per method and dataset. We model all final performances from our experiments as a parametric function of the compute budget \( P(k) \).

\textbf{Extrapolation from Parametric Fits.} \Cref{fig:parametric-fit} shows the fitted curves for each method. Since the shape of these curves is dataset and model dependent, we cannot predict exact performance for different levels of data selection compute. However, the close fit of the parametric models obtained from smaller models allows us to estimate the compute-optimal ratio between the training model size and the selection model size. For perplexity-based data selection, our extrapolation suggests that the method becomes compute-optimal when the training model is 5x larger than the data selection model—around 35B parameters. For gradient-based data selection, our extrapolation indicates that the training model needs to be approximately 10x larger than the data selection model to be compute-optimal—around 70B parameters. Appendix~\ref{Appendix-Extrapolation} presents our results and details the extrapolation.

\section{Analysis}

\begin{figure}[!t]
    \centering
    \resizebox{1.0\textwidth}{!}{%
    \includegraphics{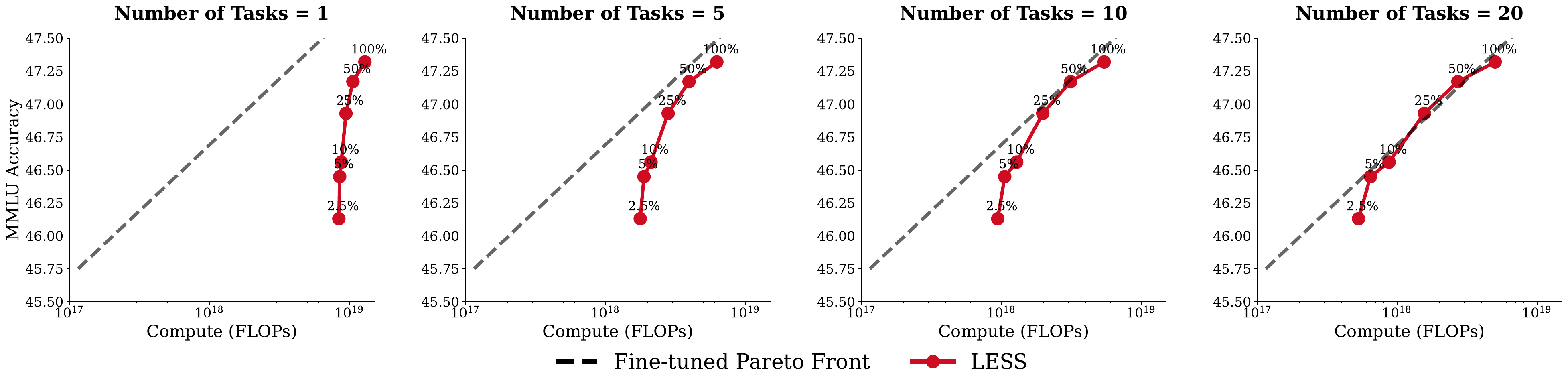} 
    }
    \caption{\textbf{Multiple Task-Specific Model Break-Even Analysis }. Costs to perform gradient-based method (LESS) are spread over all the target tasks. Performance under compute-constraints reach the finetuned Pareto frontier at 10 tasks, surpassing it at 20 tasks.}
    \label{fig:break-even-analysis}
\end{figure}

\begin{figure}[!t]
    \centering
    \subfloat[\label{fig:fixed_training_budget}]{
        \includegraphics[width=0.32\textwidth]{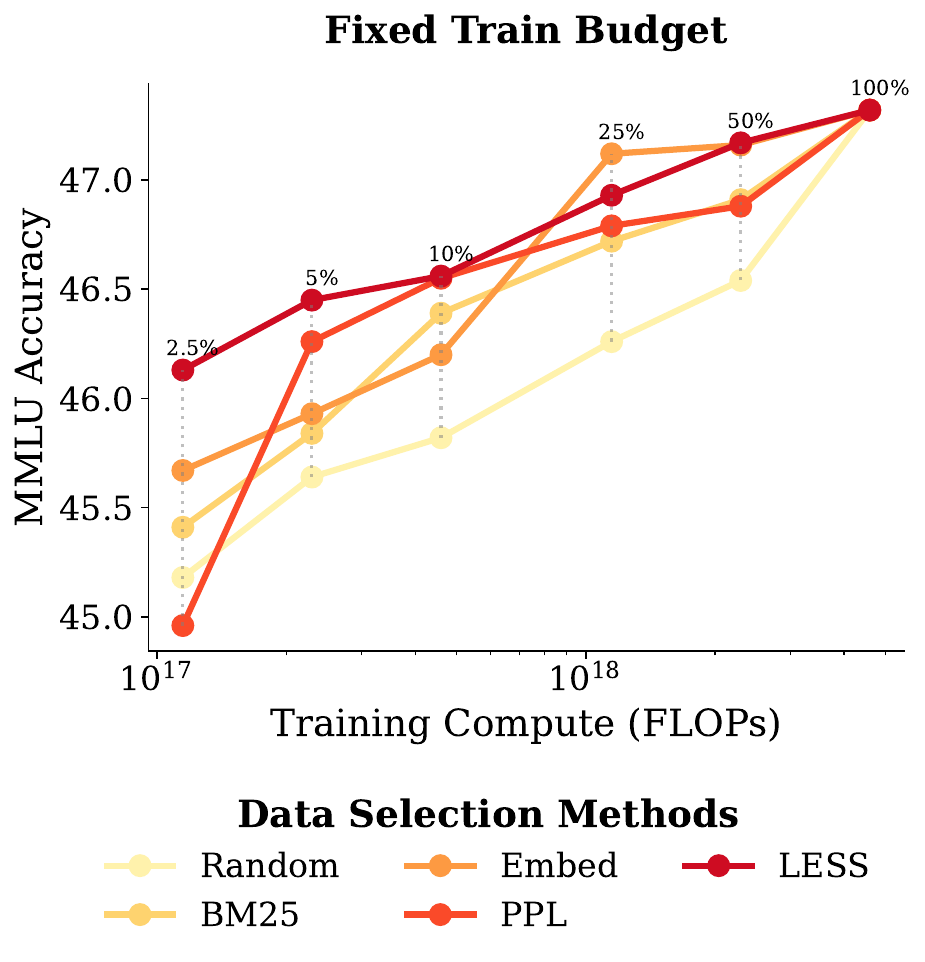}
    }
    \hfill
    \subfloat[\label{fig:IFEval_results_7B:}]{
        \includegraphics[width=0.61\textwidth]{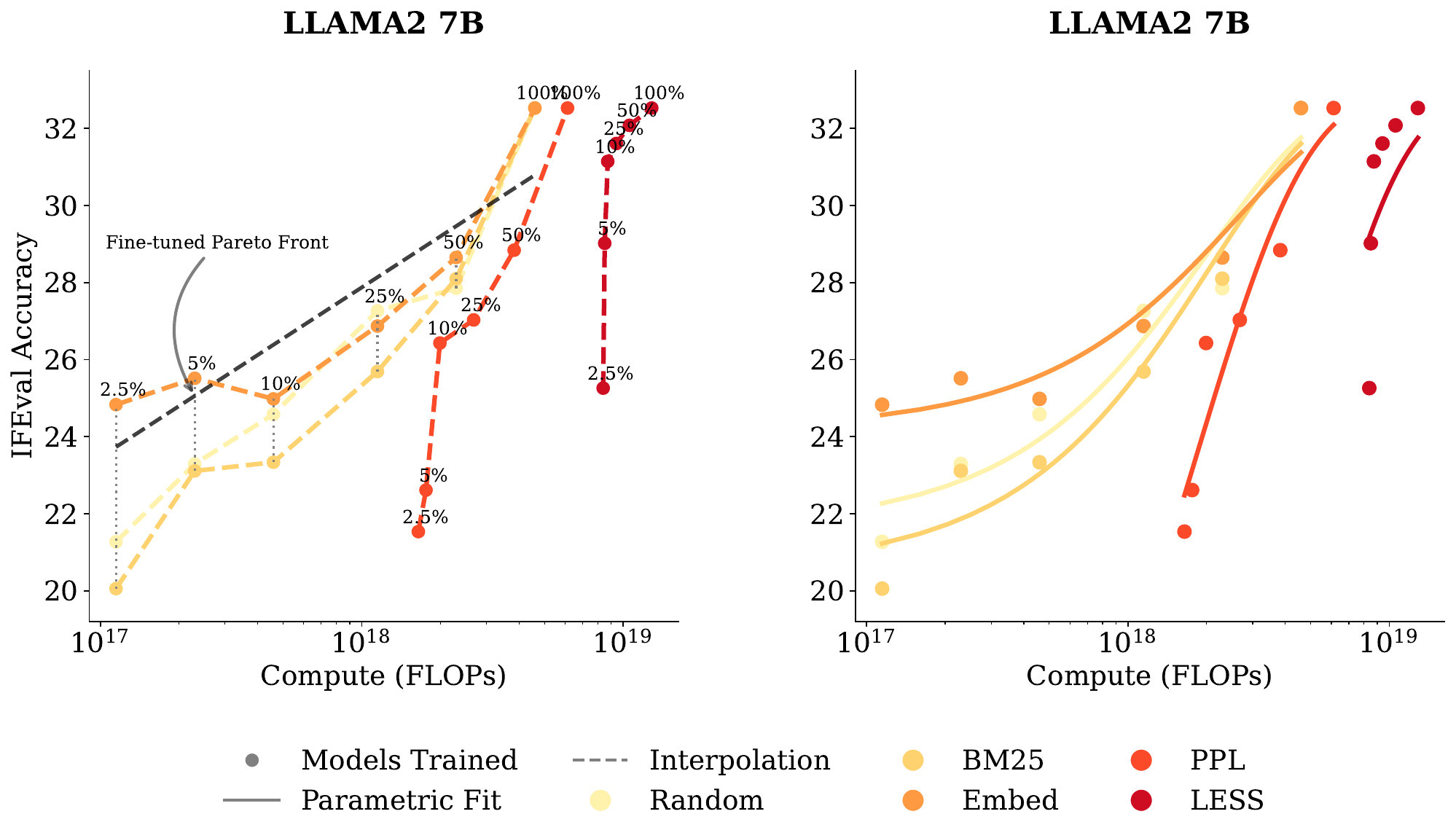}
    }
    \caption{ (a) \textbf{Fixed Training Budget.} Considering only training budget, sophisticated methods consistently outperforms cheaper methods. (b) \textbf{Performance and Parametric Fit} on IFEval. At small compute budget, sophisticated methods are not compute optimal.}
    \label{fig:results-analysis-combined}
\end{figure}

% \subsection{Efficiency of Data Selection Methods}
\paragraph{Comparing Training versus Total Compute Budgeting.} While our primary interest is in the full compute constrained setting, we note that different results hold if targeting only a small training budget as in~\Cref{eq:equation-1}. When the training-budget is fixed, we observe in ~\Cref{fig:results-analysis-combined} (a) that the gradient-based method (LESS) consistently outperforms other data selection methods, with the perplexity-based method (PPL) being the closest competitor. 

%While powerful data selection methods can reduce the amount of training data needed to achieve the same performance, the cost of data selection often outweighs the resulting training speedup.

%While powerful data selection methods can reduce the amount of training data needed to achieve the same performance, the cost of data selection often outweighs the resulting training speedup.

% \begin{figure}[t]
%     \centering
%     \begin{minipage}[t]{0.35\textwidth}
%         \centering
%         \includegraphics[width=\textwidth]{figures/Draft/MMLU_7B_Comparison_Single.pdf}
%         \caption{\textbf{Data Budget vs. Compute Budget}. }
%         \label{fig:flops-normalization}
%     \end{minipage}%
%     \hfill
%     \begin{minipage}[t]{0.61\textwidth}
%         \centering
%         \vspace{-51mm}  % Adjust this value as needed
%         \includegraphics[width=\textwidth]{figures/Draft/MMLU 8B Comparison.pdf}
%         \caption{\textbf{Performance and Parametric Fit of Performance on IFEval}. }
%         \label{fig:data-similarity}
%     \end{minipage}
% \end{figure}

\paragraph{Multiple Task-Specific Models.}
Another interesting setting is when the goal is to train
multiple models from the same large training set, where each targets a different task. In this setting, the  
gradient information calculation can be amortized between target tasks. However, even in this setting the cost of gradient calculation is still quite high. As shown in \Cref{fig:break-even-analysis}, with  \textsc{Llama}2 7B as the data selection model, LESS needs 20 tasks to surpass the finetuned pareto frontier. We perform similar analysis for other model sizes in \Cref{Appendix-Break-Even-Analysis}.

% "#of tasks to break even at 5% per task".

\begin{figure}[t!]
    \centering
    \resizebox{0.85\textwidth}{!}{%
    \includegraphics{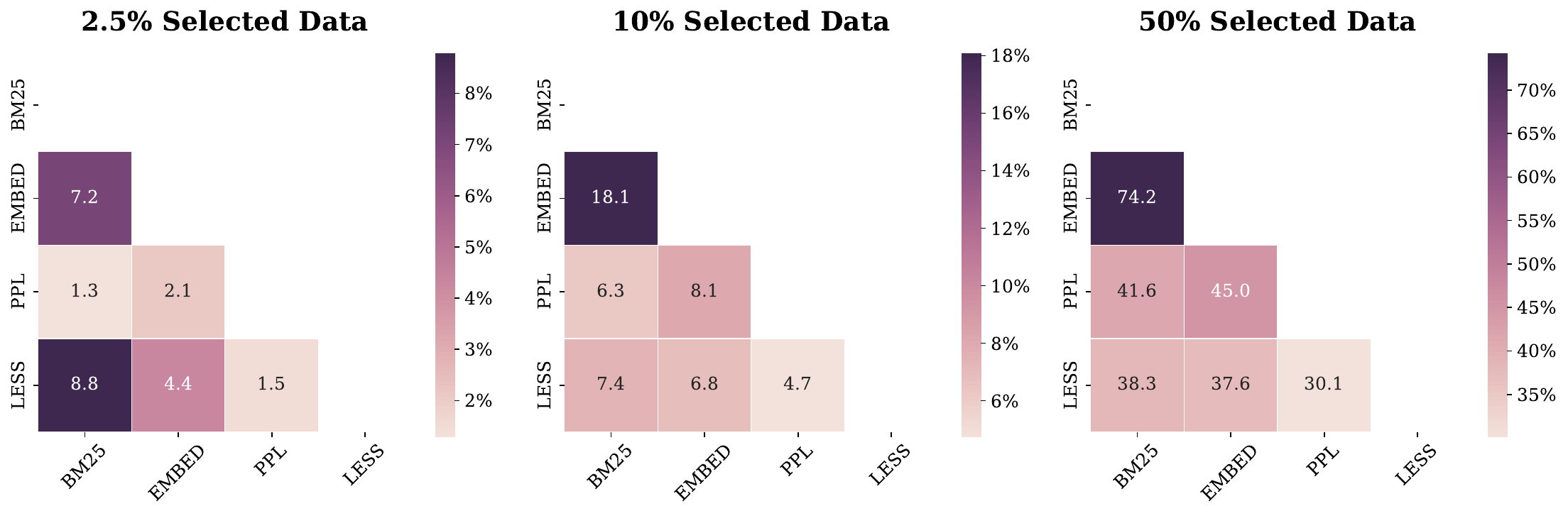} 
    }
    \caption{\textbf{Data Similarity Between Data Selection Methods} for MMLU. }
    \label{fig:data-similarity}
\end{figure}

\paragraph{Data Similarity.} \Cref{fig:data-similarity} presents the Jaccard similarity of data selected by different data selection methods. We observe that BM25 and Embed tend to select highly similar data, which explains their comparable performance. In contrast, the data selected by LESS bears the least resemblance to the other methods. We perform similar analysis for different target tasks in \Cref{Appendix-Data-Similarity-Analysis}.

% \begin{figure}
%   \centering
%    \resizebox{1.0\textwidth}{!}{%
%   \subfloat{\includegraphics{figures/Draft/MMLU DATA SIMILARITY ICLR DRAFT 1.png} \label{fig:a}}} \\
%  \resizebox{1.0\textwidth}{!}{%
%   \subfloat{\includegraphics{figures/Draft/BBH DATA SIMILARITY ICLR DRAFT 1.png} \label{fig:b}}
%   }
%   \caption{\textbf{Data Similarity.}} \label{fig:AB}
% \end{figure}

% \colorbox{blue!30}{1. Compared different data selection methods in terms of the data that they are selecting. }

% \colorbox{blue!30}{2. Show some kind of similarity score between each data selection ranking. (use some kind of linear regression?) }

% \subsection{Model Size is the Best Data Selection}

% \subsection{Sophisticated Data-Selection is currently compute inefficient}

\newpage
\section{Conclusion}
\label{Conclusion}

Data selection is a valuable tool for fine-tuning language models, but its primary benefit --- saving compute --- must be balanced against the compute costs required to identify an optimal dataset. In practice, popular methods are surprisingly compute intensive, yielding training data reduction at the cost of worse compute-constrained performance. We hope that these results motivate further research into more compute-efficient data selection methods. As demonstrated by our 70B finetuning experiments, sophisticated data selection methods can leverage smaller models for data selection, allowing larger models from the same family to be trained more efficiently in the LLM setting.

% Our analysis is limitee. Further, we only consider 1-epoch training, where in practice multiple-epochs training over selected examples are 

\section{Acknowledgments}
\label{Acknowledgements}

We would like to thank Woojeong Kim and Celine Lee for their insightful feedback on the paper’s writing. We are grateful to Mengzhou Xia for assistance with setting up the codebase and valuable discussions on data selection. Additionally, we thank Junxiong Wang,  Jing Nathan Yan, and Wenting Zhao for their support in conducting the early experiments. This work was supported by NSF IIS-1901030 and NSF CAREER 2037519.

\bibliography{iclr2025_conference}

\begin{thebibliography}{50}
\providecommand{\natexlab}[1]{#1}
\providecommand{\url}[1]{\texttt{#1}}
\expandafter\ifx\csname urlstyle\endcsname\relax
  \providecommand{\doi}[1]{doi: #1}\else
  \providecommand{\doi}{doi: \begingroup \urlstyle{rm}\Url}\fi

\bibitem[Albalak et~al.(2024)Albalak, Elazar, Xie, Longpre, Lambert, Wang, Muennighoff, Hou, Pan, Jeong, et~al.]{albalak2024survey}
Alon Albalak, Yanai Elazar, Sang~Michael Xie, Shayne Longpre, Nathan Lambert, Xinyi Wang, Niklas Muennighoff, Bairu Hou, Liangming Pan, Haewon Jeong, et~al.
\newblock A survey on data selection for language models.
\newblock \emph{arXiv preprint arXiv:2402.16827}, 2024.

\bibitem[Ankner et~al.(2024)Ankner, Blakeney, Sreenivasan, Marion, Leavitt, and Paul]{ankner2024perplexed}
Zachary Ankner, Cody Blakeney, Kartik Sreenivasan, Max Marion, Matthew~L Leavitt, and Mansheej Paul.
\newblock Perplexed by perplexity: Perplexity-based data pruning with small reference models.
\newblock \emph{arXiv preprint arXiv:2405.20541}, 2024.

\bibitem[Antonello et~al.(2020)Antonello, Beckage, Turek, and Huth]{antonello2020selecting}
Richard Antonello, Nicole Beckage, Javier Turek, and Alexander Huth.
\newblock Selecting informative contexts improves language model finetuning.
\newblock \emph{arXiv preprint arXiv:2005.00175}, 2020.

\bibitem[Bi et~al.(2024)Bi, Chen, Chen, Chen, Dai, Deng, Ding, Dong, Du, Fu, et~al.]{bi2024deepseek}
Xiao Bi, Deli Chen, Guanting Chen, Shanhuang Chen, Damai Dai, Chengqi Deng, Honghui Ding, Kai Dong, Qiushi Du, Zhe Fu, et~al.
\newblock Deepseek llm: Scaling open-source language models with longtermism.
\newblock \emph{arXiv preprint arXiv:2401.02954}, 2024.

\bibitem[Chen et~al.(2023)Chen, Li, Yan, Wang, Gunaratna, Yadav, Tang, Srinivasan, Zhou, Huang, et~al.]{chen2023alpagasus}
Lichang Chen, Shiyang Li, Jun Yan, Hai Wang, Kalpa Gunaratna, Vikas Yadav, Zheng Tang, Vijay Srinivasan, Tianyi Zhou, Heng Huang, et~al.
\newblock Alpagasus: Training a better alpaca with fewer data.
\newblock \emph{arXiv preprint arXiv:2307.08701}, 2023.

\bibitem[Chen et~al.(2024)Chen, Roberts, Bhatia, Wang, Zhang, Sala, and R{\'e}]{chen2024skill}
Mayee Chen, Nicholas Roberts, Kush Bhatia, Jue Wang, Ce~Zhang, Frederic Sala, and Christopher R{\'e}.
\newblock Skill-it! a data-driven skills framework for understanding and training language models.
\newblock \emph{Advances in Neural Information Processing Systems}, 36, 2024.

\bibitem[Conover et~al.(2023)Conover, Hayes, Mathur, Xie, Wan, Shah, Ghodsi, Wendell, Zaharia, and Xin]{conover2023free}
Mike Conover, Matt Hayes, Ankit Mathur, Jianwei Xie, Jun Wan, Sam Shah, Ali Ghodsi, Patrick Wendell, Matei Zaharia, and Reynold Xin.
\newblock Free dolly: Introducing the world’s first truly open instruction-tuned llm.
\newblock \emph{Company Blog of Databricks}, 2023.

\bibitem[Dubey et~al.(2024)Dubey, Jauhri, Pandey, Kadian, Al-Dahle, Letman, Mathur, Schelten, Yang, Fan, et~al.]{dubey2024llama}
Abhimanyu Dubey, Abhinav Jauhri, Abhinav Pandey, Abhishek Kadian, Ahmad Al-Dahle, Aiesha Letman, Akhil Mathur, Alan Schelten, Amy Yang, Angela Fan, et~al.
\newblock The llama 3 herd of models.
\newblock \emph{arXiv preprint arXiv:2407.21783}, 2024.

\bibitem[Goyal et~al.(2024)Goyal, Maini, Lipton, Raghunathan, and Kolter]{goyal2024scaling}
Sachin Goyal, Pratyush Maini, Zachary~C Lipton, Aditi Raghunathan, and J~Zico Kolter.
\newblock Scaling laws for data filtering--data curation cannot be compute agnostic.
\newblock In \emph{Proceedings of the IEEE/CVF Conference on Computer Vision and Pattern Recognition}, pp.\  22702--22711, 2024.

\bibitem[Han(2023)]{han2023context}
Xiaochuang Han.
\newblock In-context alignment: Chat with vanilla language models before fine-tuning.
\newblock \emph{arXiv preprint arXiv:2308.04275}, 2023.

\bibitem[Hart(1968)]{hart1968condensed}
Peter Hart.
\newblock The condensed nearest neighbor rule (corresp.).
\newblock \emph{IEEE transactions on information theory}, 14\penalty0 (3):\penalty0 515--516, 1968.

\bibitem[Hendrycks et~al.(2020)Hendrycks, Burns, Basart, Zou, Mazeika, Song, and Steinhardt]{hendrycks2020measuring}
Dan Hendrycks, Collin Burns, Steven Basart, Andy Zou, Mantas Mazeika, Dawn Song, and Jacob Steinhardt.
\newblock Measuring massive multitask language understanding.
\newblock \emph{arXiv preprint arXiv:2009.03300}, 2020.

\bibitem[Hernandez et~al.(2021)Hernandez, Kaplan, Henighan, and McCandlish]{hernandez2021scaling}
Danny Hernandez, Jared Kaplan, Tom Henighan, and Sam McCandlish.
\newblock Scaling laws for transfer.
\newblock \emph{arXiv preprint arXiv:2102.01293}, 2021.

\bibitem[Hoffmann et~al.(2022)Hoffmann, Borgeaud, Mensch, Buchatskaya, Cai, Rutherford, Casas, Hendricks, Welbl, Clark, et~al.]{hoffmann2022training}
Jordan Hoffmann, Sebastian Borgeaud, Arthur Mensch, Elena Buchatskaya, Trevor Cai, Eliza Rutherford, Diego de~Las Casas, Lisa~Anne Hendricks, Johannes Welbl, Aidan Clark, et~al.
\newblock Training compute-optimal large language models.
\newblock \emph{arXiv preprint arXiv:2203.15556}, 2022.

\bibitem[Hu et~al.(2021)Hu, Shen, Wallis, Allen-Zhu, Li, Wang, Wang, and Chen]{hu2021lora}
Edward~J Hu, Yelong Shen, Phillip Wallis, Zeyuan Allen-Zhu, Yuanzhi Li, Shean Wang, Lu~Wang, and Weizhu Chen.
\newblock Lora: Low-rank adaptation of large language models.
\newblock \emph{arXiv preprint arXiv:2106.09685}, 2021.

\bibitem[Isik et~al.(2024)Isik, Ponomareva, Hazimeh, Paparas, Vassilvitskii, and Koyejo]{isik2024scaling}
Berivan Isik, Natalia Ponomareva, Hussein Hazimeh, Dimitris Paparas, Sergei Vassilvitskii, and Sanmi Koyejo.
\newblock Scaling laws for downstream task performance of large language models.
\newblock \emph{arXiv preprint arXiv:2402.04177}, 2024.

\bibitem[Ivison et~al.(2022)Ivison, Smith, Hajishirzi, and Dasigi]{ivison2022data}
Hamish Ivison, Noah~A Smith, Hannaneh Hajishirzi, and Pradeep Dasigi.
\newblock Data-efficient finetuning using cross-task nearest neighbors.
\newblock \emph{arXiv preprint arXiv:2212.00196}, 2022.

\bibitem[Ivison et~al.(2023)Ivison, Wang, Pyatkin, Lambert, Peters, Dasigi, Jang, Wadden, Smith, Beltagy, et~al.]{ivison2023camels}
Hamish Ivison, Yizhong Wang, Valentina Pyatkin, Nathan Lambert, Matthew Peters, Pradeep Dasigi, Joel Jang, David Wadden, Noah~A Smith, Iz~Beltagy, et~al.
\newblock Camels in a changing climate: Enhancing lm adaptation with tulu 2.
\newblock \emph{arXiv preprint arXiv:2311.10702}, 2023.

\bibitem[John(1975)]{john1975d}
RC~St John.
\newblock D-optimality for regression designs: a review.
\newblock \emph{Technometrics}, 17\penalty0 (1):\penalty0 15--23, 1975.

\bibitem[Kaplan et~al.(2020)Kaplan, McCandlish, Henighan, Brown, Chess, Child, Gray, Radford, Wu, and Amodei]{kaplan2020scaling}
Jared Kaplan, Sam McCandlish, Tom Henighan, Tom~B Brown, Benjamin Chess, Rewon Child, Scott Gray, Alec Radford, Jeffrey Wu, and Dario Amodei.
\newblock Scaling laws for neural language models.
\newblock \emph{arXiv preprint arXiv:2001.08361}, 2020.

\bibitem[Killamsetty et~al.(2021{\natexlab{a}})Killamsetty, Durga, Ramakrishnan, De, and Iyer]{killamsetty2021grad}
Krishnateja Killamsetty, Sivasubramanian Durga, Ganesh Ramakrishnan, Abir De, and Rishabh Iyer.
\newblock Grad-match: Gradient matching based data subset selection for efficient deep model training.
\newblock In \emph{International Conference on Machine Learning}, pp.\  5464--5474. PMLR, 2021{\natexlab{a}}.

\bibitem[Killamsetty et~al.(2021{\natexlab{b}})Killamsetty, Sivasubramanian, Ramakrishnan, and Iyer]{killamsetty2021glister}
Krishnateja Killamsetty, Durga Sivasubramanian, Ganesh Ramakrishnan, and Rishabh Iyer.
\newblock Glister: Generalization based data subset selection for efficient and robust learning.
\newblock In \emph{Proceedings of the AAAI Conference on Artificial Intelligence}, volume~35, pp.\  8110--8118, 2021{\natexlab{b}}.

\bibitem[Kirchhoff \& Bilmes(2014)Kirchhoff and Bilmes]{kirchhoff2014submodularity}
Katrin Kirchhoff and Jeff Bilmes.
\newblock Submodularity for data selection in machine translation.
\newblock In \emph{Proceedings of the 2014 Conference on Empirical Methods in Natural Language Processing (EMNLP)}, pp.\  131--141, 2014.

\bibitem[K{\"o}pf et~al.(2024)K{\"o}pf, Kilcher, von R{\"u}tte, Anagnostidis, Tam, Stevens, Barhoum, Nguyen, Stanley, Nagyfi, et~al.]{kopf2024openassistant}
Andreas K{\"o}pf, Yannic Kilcher, Dimitri von R{\"u}tte, Sotiris Anagnostidis, Zhi~Rui Tam, Keith Stevens, Abdullah Barhoum, Duc Nguyen, Oliver Stanley, Rich{\'a}rd Nagyfi, et~al.
\newblock Openassistant conversations-democratizing large language model alignment.
\newblock \emph{Advances in Neural Information Processing Systems}, 36, 2024.

\bibitem[Lin et~al.(2024)Lin, Huang, Ye, Chen, Wang, Li, Ma, Wan, Zou, and Liang]{lin2024selecting}
Haowei Lin, Baizhou Huang, Haotian Ye, Qinyu Chen, Zihao Wang, Sujian Li, Jianzhu Ma, Xiaojun Wan, James Zou, and Yitao Liang.
\newblock Selecting large language model to fine-tune via rectified scaling law.
\newblock \emph{arXiv preprint arXiv:2402.02314}, 2024.

\bibitem[Liu et~al.(2021)Liu, Shen, Zhang, Dolan, Carin, and Chen]{liu2021makes}
Jiachang Liu, Dinghan Shen, Yizhe Zhang, Bill Dolan, Lawrence Carin, and Weizhu Chen.
\newblock What makes good in-context examples for gpt-$3 $?
\newblock \emph{arXiv preprint arXiv:2101.06804}, 2021.

\bibitem[Longpre et~al.(2023)Longpre, Hou, Vu, Webson, Chung, Tay, Zhou, Le, Zoph, Wei, et~al.]{longpre2023flan}
Shayne Longpre, Le~Hou, Tu~Vu, Albert Webson, Hyung~Won Chung, Yi~Tay, Denny Zhou, Quoc~V Le, Barret Zoph, Jason Wei, et~al.
\newblock The flan collection: Designing data and methods for effective instruction tuning.
\newblock In \emph{International Conference on Machine Learning}, pp.\  22631--22648. PMLR, 2023.

\bibitem[Lu et~al.(2023)Lu, Yuan, Yuan, Lin, Lin, Tan, Zhou, and Zhou]{lu2023instag}
Keming Lu, Hongyi Yuan, Zheng Yuan, Runji Lin, Junyang Lin, Chuanqi Tan, Chang Zhou, and Jingren Zhou.
\newblock \# instag: Instruction tagging for analyzing supervised fine-tuning of large language models.
\newblock In \emph{The Twelfth International Conference on Learning Representations}, 2023.

\bibitem[Marion et~al.(2023)Marion, {\"U}st{\"u}n, Pozzobon, Wang, Fadaee, and Hooker]{marion2023less}
Max Marion, Ahmet {\"U}st{\"u}n, Luiza Pozzobon, Alex Wang, Marzieh Fadaee, and Sara Hooker.
\newblock When less is more: Investigating data pruning for pretraining llms at scale.
\newblock \emph{arXiv preprint arXiv:2309.04564}, 2023.

\bibitem[Mirzasoleiman et~al.(2020)Mirzasoleiman, Bilmes, and Leskovec]{mirzasoleiman2020coresets}
Baharan Mirzasoleiman, Jeff Bilmes, and Jure Leskovec.
\newblock Coresets for data-efficient training of machine learning models.
\newblock In \emph{International Conference on Machine Learning}, pp.\  6950--6960. PMLR, 2020.

\bibitem[Mishra et~al.(2021)Mishra, Khashabi, Baral, and Hajishirzi]{mishra2021cross}
Swaroop Mishra, Daniel Khashabi, Chitta Baral, and Hannaneh Hajishirzi.
\newblock Cross-task generalization via natural language crowdsourcing instructions.
\newblock \emph{arXiv preprint arXiv:2104.08773}, 2021.

\bibitem[Muennighoff et~al.(2024)Muennighoff, Rush, Barak, Le~Scao, Tazi, Piktus, Pyysalo, Wolf, and Raffel]{muennighoff2024scaling}
Niklas Muennighoff, Alexander Rush, Boaz Barak, Teven Le~Scao, Nouamane Tazi, Aleksandra Piktus, Sampo Pyysalo, Thomas Wolf, and Colin~A Raffel.
\newblock Scaling data-constrained language models.
\newblock \emph{Advances in Neural Information Processing Systems}, 36, 2024.

\bibitem[Ni et~al.(2021)Ni, Qu, Lu, Dai, {\'A}brego, Ma, Zhao, Luan, Hall, Chang, et~al.]{ni2021large}
Jianmo Ni, Chen Qu, Jing Lu, Zhuyun Dai, Gustavo~Hern{\'a}ndez {\'A}brego, Ji~Ma, Vincent~Y Zhao, Yi~Luan, Keith~B Hall, Ming-Wei Chang, et~al.
\newblock Large dual encoders are generalizable retrievers.
\newblock \emph{arXiv preprint arXiv:2112.07899}, 2021.

\bibitem[Pruthi et~al.(2020)Pruthi, Liu, Kale, and Sundararajan]{pruthi2020estimating}
Garima Pruthi, Frederick Liu, Satyen Kale, and Mukund Sundararajan.
\newblock Estimating training data influence by tracing gradient descent.
\newblock \emph{Advances in Neural Information Processing Systems}, 33:\penalty0 19920--19930, 2020.

\bibitem[Robertson et~al.(2009)Robertson, Zaragoza, et~al.]{robertson2009probabilistic}
Stephen Robertson, Hugo Zaragoza, et~al.
\newblock The probabilistic relevance framework: Bm25 and beyond.
\newblock \emph{Foundations and Trends{\textregistered} in Information Retrieval}, 3\penalty0 (4):\penalty0 333--389, 2009.

\bibitem[Rubin et~al.(2021)Rubin, Herzig, and Berant]{rubin2021learning}
Ohad Rubin, Jonathan Herzig, and Jonathan Berant.
\newblock Learning to retrieve prompts for in-context learning.
\newblock \emph{arXiv preprint arXiv:2112.08633}, 2021.

\bibitem[Sanh et~al.(2021)Sanh, Webson, Raffel, Bach, Sutawika, Alyafeai, Chaffin, Stiegler, Scao, Raja, et~al.]{sanh2021multitask}
Victor Sanh, Albert Webson, Colin Raffel, Stephen~H Bach, Lintang Sutawika, Zaid Alyafeai, Antoine Chaffin, Arnaud Stiegler, Teven~Le Scao, Arun Raja, et~al.
\newblock Multitask prompted training enables zero-shot task generalization.
\newblock \emph{arXiv preprint arXiv:2110.08207}, 2021.

\bibitem[Silva \& Barbosa(2024)Silva and Barbosa]{silva2024improving}
Levy Silva and Luciano Barbosa.
\newblock Improving dense retrieval models with llm augmented data for dataset search.
\newblock \emph{Knowledge-Based Systems}, 294:\penalty0 111740, 2024.

\bibitem[Srivastava et~al.(2022)Srivastava, Rastogi, Rao, Shoeb, Abid, Fisch, Brown, Santoro, Gupta, Garriga-Alonso, et~al.]{srivastava2022beyond}
Aarohi Srivastava, Abhinav Rastogi, Abhishek Rao, Abu Awal~Md Shoeb, Abubakar Abid, Adam Fisch, Adam~R Brown, Adam Santoro, Aditya Gupta, Adri{\`a} Garriga-Alonso, et~al.
\newblock Beyond the imitation game: Quantifying and extrapolating the capabilities of language models.
\newblock \emph{arXiv preprint arXiv:2206.04615}, 2022.

\bibitem[Suzgun et~al.(2022)Suzgun, Scales, Sch{\"a}rli, Gehrmann, Tay, Chung, Chowdhery, Le, Chi, Zhou, et~al.]{suzgun2022challenging}
Mirac Suzgun, Nathan Scales, Nathanael Sch{\"a}rli, Sebastian Gehrmann, Yi~Tay, Hyung~Won Chung, Aakanksha Chowdhery, Quoc~V Le, Ed~H Chi, Denny Zhou, et~al.
\newblock Challenging big-bench tasks and whether chain-of-thought can solve them.
\newblock \emph{arXiv preprint arXiv:2210.09261}, 2022.

\bibitem[Touvron et~al.(2023)Touvron, Martin, Stone, Albert, Almahairi, Babaei, Bashlykov, Batra, Bhargava, Bhosale, et~al.]{touvron2023llama}
Hugo Touvron, Louis Martin, Kevin Stone, Peter Albert, Amjad Almahairi, Yasmine Babaei, Nikolay Bashlykov, Soumya Batra, Prajjwal Bhargava, Shruti Bhosale, et~al.
\newblock Llama 2: Open foundation and fine-tuned chat models.
\newblock \emph{arXiv preprint arXiv:2307.09288}, 2023.

\bibitem[Wang et~al.(2023)Wang, Ivison, Dasigi, Hessel, Khot, Chandu, Wadden, MacMillan, Smith, Beltagy, et~al.]{wang2023far}
Yizhong Wang, Hamish Ivison, Pradeep Dasigi, Jack Hessel, Tushar Khot, Khyathi Chandu, David Wadden, Kelsey MacMillan, Noah~A Smith, Iz~Beltagy, et~al.
\newblock How far can camels go? exploring the state of instruction tuning on open resources.
\newblock \emph{Advances in Neural Information Processing Systems}, 36:\penalty0 74764--74786, 2023.

\bibitem[Wei et~al.(2021)Wei, Bosma, Zhao, Guu, Yu, Lester, Du, Dai, and Le]{wei2021finetuned}
Jason Wei, Maarten Bosma, Vincent~Y Zhao, Kelvin Guu, Adams~Wei Yu, Brian Lester, Nan Du, Andrew~M Dai, and Quoc~V Le.
\newblock Finetuned language models are zero-shot learners.
\newblock \emph{arXiv preprint arXiv:2109.01652}, 2021.

\bibitem[Wei et~al.(2022)Wei, Wang, Schuurmans, Bosma, Xia, Chi, Le, Zhou, et~al.]{wei2022chain}
Jason Wei, Xuezhi Wang, Dale Schuurmans, Maarten Bosma, Fei Xia, Ed~Chi, Quoc~V Le, Denny Zhou, et~al.
\newblock Chain-of-thought prompting elicits reasoning in large language models.
\newblock \emph{Advances in neural information processing systems}, 35:\penalty0 24824--24837, 2022.

\bibitem[Wettig et~al.(2024)Wettig, Gupta, Malik, and Chen]{wettig2024qurating}
Alexander Wettig, Aatmik Gupta, Saumya Malik, and Danqi Chen.
\newblock Qurating: Selecting high-quality data for training language models.
\newblock \emph{arXiv preprint arXiv:2402.09739}, 2024.

\bibitem[Xia et~al.(2024)Xia, Malladi, Gururangan, Arora, and Chen]{xia2024less}
Mengzhou Xia, Sadhika Malladi, Suchin Gururangan, Sanjeev Arora, and Danqi Chen.
\newblock Less: Selecting influential data for targeted instruction tuning.
\newblock \emph{arXiv preprint arXiv:2402.04333}, 2024.

\bibitem[Xie et~al.(2023)Xie, Santurkar, Ma, and Liang]{xie2023data}
Sang~Michael Xie, Shibani Santurkar, Tengyu Ma, and Percy~S Liang.
\newblock Data selection for language models via importance resampling.
\newblock \emph{Advances in Neural Information Processing Systems}, 36:\penalty0 34201--34227, 2023.

\bibitem[Zhang et~al.(2024)Zhang, Liu, Cherry, and Firat]{zhang2024scaling}
Biao Zhang, Zhongtao Liu, Colin Cherry, and Orhan Firat.
\newblock When scaling meets llm finetuning: The effect of data, model and finetuning method.
\newblock \emph{arXiv preprint arXiv:2402.17193}, 2024.

\bibitem[Zhou et~al.(2024)Zhou, Liu, Xu, Iyer, Sun, Mao, Ma, Efrat, Yu, Yu, et~al.]{zhou2024lima}
Chunting Zhou, Pengfei Liu, Puxin Xu, Srinivasan Iyer, Jiao Sun, Yuning Mao, Xuezhe Ma, Avia Efrat, Ping Yu, Lili Yu, et~al.
\newblock Lima: Less is more for alignment.
\newblock \emph{Advances in Neural Information Processing Systems}, 36, 2024.

\bibitem[Zhou et~al.(2023)Zhou, Lu, Mishra, Brahma, Basu, Luan, Zhou, and Hou]{zhou2023instruction}
Jeffrey Zhou, Tianjian Lu, Swaroop Mishra, Siddhartha Brahma, Sujoy Basu, Yi~Luan, Denny Zhou, and Le~Hou.
\newblock Instruction-following evaluation for large language models.
\newblock \emph{arXiv preprint arXiv:2311.07911}, 2023.

\end{thebibliography}
\bibliographystyle{iclr2025_conference}

\newpage
\appendix
\section{FLOPs Computation}
To compute the training FLOPs of a transformer model, we assume the backward pass has twice the FLOPs of the forward pass and follow the FLOP counts of various components of a transformer model for a single forward pass as in \citet{kaplan2020scaling} as in \Cref{tab:operation_flops}. We account for all training FLOPs in our analysis, including those from the embedding matrices. For large models, the contribution of FLOPs and parameters from embedding matrices is minimal. We apply a factor of 2 to represent the multiply-accumulate cost (MAC).

\begin{table}[h]
\centering
\resizebox{\textwidth}{!}{%
\begin{tabular}{>{\raggedright}m{4cm} >{\raggedright}m{5cm} >{\raggedright\arraybackslash}m{5cm}}
\toprule
\textbf{Operation} & \textbf{Parameters} & \textbf{FLOPs per Token} \\ 
\midrule
Embed & $n_{\text{vocab}}d_{\text{model}}$ & $2n_{\text{vocab}}d_{\text{model}}$ \\ 
\midrule
Attention: QKV & $n_{\text{layer}}d_{\text{model}}3d_{\text{attn}}$ & $2n_{\text{layer}}d_{\text{model}}3d_{\text{attn}}$ \\ 
Attention: Mask (Q-K Dot) & \textemdash & $2n_{\text{layer}}n_{\text{ctx}}d_{\text{attn}}$ \\ 
% Attention: Softmax & \textemdash  & $3n_{\text{layer}}n_{\text{ctx}}n_{\text{head}}$ \\ 
% Attention: Softmax@Query Reduction & \textemdash & $2n_{\text{layer}}n_{\text{ctx}}n_{\text{head}}$ \\ 
Attention: Project & $n_{\text{layer}}d_{\text{attn}}d_{\text{model}}$ & $2n_{\text{layer}}d_{\text{attn}}d_{\text{model}}$ \\ 
\midrule
Feedforward & $n_{\text{layer}}3d_{\text{model}}d_{\text{ff}}$ & $2n_{\text{layer}}3d_{\text{model}}d_{\text{ff}}$ \\ 
\midrule
De-embed (Head) & $d_{\text{model}}n_{\text{vocab}}$ & $2d_{\text{model}}n_{\text{vocab}}$ \\ 
\midrule
Total (Non-Embedding) & $N^{\prime} = 2d_{\text{model}}n_{\text{layer}}(2d_{\text{attn}} + 1.5d_{\text{ff}}) + d_{\text{model}}n_{\text{vocab}}$ & $C_{\text{forward}} = 2N^{\prime} + 2n_{\text{layer}}n_{\text{ctx}}d_{\text{attn}}$ \\ 
\midrule
Total & $N = 2d_{\text{model}}n_{\text{layer}}(2d_{\text{attn}} + 1.5d_{\text{ff}}) + d_{\text{model}}n_{\text{vocab}} + n_{\text{vocab}}d_{\text{model}}$ & $C_{\text{forward}} = 2N + 2n_{\text{layer}}n_{\text{ctx}}d_{\text{attn}}$  \\ 
\bottomrule
\end{tabular}
}
\caption{Parameters and FLOPs per token for different operations.}
\label{tab:operation_flops}
\end{table}

 For better granularity, we show a comparision between our calculation and one that uses the common approximation $C_{train}=6ND$, where $C_{train}$ denotes FLOPs, $D$ denotes total number of training tokens, and $N$ is the number of parameters in \Cref{tab:flop_comparison}. In practically, the differences in FLOP calculation is small and thus pose no difference our analysis. 

\begin{table}[h]
\centering
\resizebox{\textwidth}{!}{
\begin{tabular}{ccccccccc}
\toprule
\textbf{Model} & $N$  & $n_{\text{layer}}$ & $n_{\text{ctx}}$ & $d_{\text{model}}$ & $d_{\text{ff}}$ & $d_{\text{attn}}$ & \textbf{FLOPs per token} & \textbf{FLOP Ratio (Ours/6ND)} \\ 
\midrule
\textsc{Llama} 2 7B & 7B & 32 & 4096 & 4096 & 11008 & $128 \times 32$  & 4.69e+10 & 1.12 \\ 
\textsc{Llama} 2 13B  & 13B & 40 & 4096 & 5120 & 13824 & $128  \times 40$ & 8.82e+10 & 1.13 \\ 
\textsc{Llama} 2 70B  & 70B & 80 & 4096 & 8192 & 28672 & $128 \times 80$ & 5.03e+11 & 1.20 \\ 
\bottomrule
\end{tabular}
}
\caption{FLOP comparison. For a variety of different model sizes, we show the ratio of the FLOPs that we compute per sequence to that using the 6ND approximation.}
\label{tab:flop_comparison}
\end{table}

 We use this in our calculation of data selection costs for methods that utilizes LLMs (i.e. perplexity-based and gradient-based).

\section{Data-Selection FLOPs}
\label{Appendix-DS-FLOPS}

In this section, we detailed our estimation of the costs to perform different data selection methods. 

\begin{table}[b]
\centering
\begin{tabular}{lc}
\toprule
\textbf{Data Selection Method} & \textbf{FLOPs Cost} \\
\midrule
BM25    & \(1 \times 10^8\) \\
Embed   & \(4.4 \times 10^{16}\) \\
PPL     & \(1.53 \times 10^{18}\) \\
LESS    & \(8.27 \times 10^{18}\) \\
\bottomrule
\end{tabular}
\caption{Data Selection Cost Summary for Each Data Selection Method}
\end{table}

\subsection{BM25}

We assume that each data point incurs a computational cost of 1 FLOP for BM25, denoted as \( c_{\text{BM25}} = 1 \). The total computation cost is approximated as proportional to the size of the dataset, \( |\mathcal{D}| \). Therefore, the data selection cost for BM25 is estimated to be \( 1 \times 10^8 \) FLOPs.

\subsection{Embed}

For the embedding-based method, we approximate the computation cost using the formula \( C_\text{forward} = 2ND \), where \( N \) represents the number of model parameters and \( D \) is the dataset size. Given that the embedding model used has \( N = 220M \) parameters \citep{ni2021large}, the data selection cost for Embed is estimated to be \( 4.4 \times 10^{16} \) FLOPs.

\subsection{PPL}

Perplexity-based methods require passing every data point through the language model. Given that the cost per token for a 7B model is \( c = 4.69 \times 10^{10} \), as shown in \Cref{tab:flop_comparison}, the data selection cost for PPL is equivalent to one forward pass. Therefore, using the \textsc{Llama}2 model for data selection, we approximate the perplexity computation cost to be \( 1.53 \times 10^{18} \) FLOPs.

\subsection{LESS}

LESS involves a two-step process: a 4-epoch warm-up training on 5\% of the dataset \( \mathcal{D} \) followed by gradient feature computation over the entire training and validation datasets. First, we calculate the FLOPs required for the 4-epoch warm-up training. Then, using the relationship between the time required for warm-up training and gradient feature computation, as outlined in \Cref{tab:less-computation-cost} from \citep{xia2024less}, we estimate the FLOPs needed for gradient feature computation. Combining these, we approximate the total cost of LESS using \textsc{Llama}2 as the data selection model to be \( 8.27 \times 10^{18} \) FLOPs.

\begin{table}[!t]
\centering
\resizebox{\textwidth}{!}{
\begin{tabular}{lcc|cc|cc}
\toprule
\textbf{} & \multicolumn{2}{c}{\textbf{Warmup LoRA Training}} & \multicolumn{2}{c}{\textbf{Gradient Features Computation}} & \multicolumn{2}{c}{\textbf{Data Selection}} \\
\cmidrule(lr){2-3} \cmidrule(lr){4-5} \cmidrule(lr){6-7}
\textbf{Compute} & \textbf{Complexity} & \textbf{Actual} & \textbf{Complexity} & \textbf{Actual} & \textbf{Complexity} & \textbf{Actual} \\
\midrule
Compute & $\mathcal{O}(|\mathcal{D}_{\text{warmup}}| \cdot N)$ & 6 Hours & $\mathcal{O}(|\mathcal{D}| \cdot N)$ & 48 Hours & $\mathcal{O}(|\mathcal{D}| \cdot |\mathcal{D}_{\text{val}}| \cdot d)$ & $\leq 1$ Min \\
Storage & - & - & $\mathcal{O}(|\mathcal{D}| \cdot N \cdot d)$ & 17.7 GB & - & - \\
\bottomrule
\end{tabular}
}
\caption{Asymptotic complexity, wall-clock runtime (measured as single A100 GPU hours), and storage cost associated with each step in LESS \cite{xia2024less}.}
\label{tab:less-computation-cost}
\end{table}

% \colorbox{orange}{TODO: Show detailed, numerical analysis of the FLOP cost of each data selection method.}

% \colorbox{yellow}{Actually just a table will do.}

\section{Parametric Function}

\subsection{Fitting Parametric Function}
\label{Appendix-Parametric-Fit}

In this appendix, we describe the process for fitting a parametric model that captures the relationship between the number of data points \( k \) and performance \( P(k) \), as a function of computational cost. The model captures diminishing returns, dependence on computational cost, and convergence toward an upper bound.

We model the expected performance \( P(k) \) after training on \( k \) data points as follows:

\[
P(k) = (\bar{P} - P_0)\times \left( 1 - \exp\left( -\lambda \frac{C(k)}{C(|\mathcal{D}|)} \right) \right) + P_0
\]

where:

\begin{itemize}
    \item \( P_0 \) is the zero-shot performance (i.e., performance without additional training).
    \item \( \bar{P} \) is the upper bound on performance (i.e., the maximum achievable performance).
    \item \( \lambda \) is a parameter controlling how efficiently the method extracts value from additional compute.
    \item \( C(k) \) is the computational cost of selecting and training on \( k \) data points.
    \item \( C(|\mathcal{D}|) \) is the total computational cost of training on the entire dataset.
\end{itemize}

The goal is to fit the parameters \( P_0 \), \( \bar{P} \), and \( \lambda \) to observed data. We fit the model by minimizing the difference between the predicted performance \( P(k) \) and the observed performance \( P_{\text{obs}}(k) \). This is formulated as the following optimization problem:

\[
\min_{P_0, \bar{P}, \lambda} \sum_{i=1}^{N} \left( P(k_i; P_0, \bar{P}, \lambda) - P_{\text{obs}, i} \right)^2
\]

where:

\begin{itemize}
    \item \( N \) is the number of data points.
    \item \( P_{\text{obs}, i} \) is the observed performance at the \( i \)-th data point.
    \item \( k_i \) is the number of data points used for training in the \( i \)-th observation.
    \item \( P(k_i; P_0, \bar{P}, \lambda) \) is the predicted performance using the parametric model.
\end{itemize}

To ensure meaningful parameter estimates, we impose the following constraints:

\begin{itemize}
    \item \( P_0 \geq 0 \), as performance cannot be negative.
    \item \( P_0 \leq \bar{P} \), ensuring that performance does not exceed the upper bound \( \bar{P} \).
    \item \( \lambda \geq 0 \), as \( \lambda \) represents the rate at which performance improves with compute.
\end{itemize}

The parameter \( \bar{P} \) is set slightly above the maximum observed performance:

\[
\bar{P} = \max_i P_{\text{obs}, i} + \epsilon
\]

where \( \epsilon = 0.05 \) is a small buffer to ensure convergence to the upper bound.

We set the initial guesses for the parameters as follows:

\begin{itemize}
    \item \( P_0^{(0)} = P_{\text{obs}, 1} \), the observed performance at zero-shot (i.e., without training).
    \item \( \bar{P}^{(0)} = \max_i P_{\text{obs}, i} \).
    \item \( \lambda^{(0)} = 1.0 \), a reasonable initial guess for the compute extraction efficiency.
\end{itemize}

We use the Levenberg-Marquardt algorithm to minimize the objective function. This method is effective for solving non-linear least squares problems and efficiently handles the non-linear nature of our parametric model.

The optimization problem is solved as follows:

\[
(P_0^\ast, \bar{P}^\ast, \lambda^\ast) = \arg\min_{P_0, \bar{P}, \lambda} \sum_{i=1}^{N} \left( P(k_i; P_0, \bar{P}, \lambda) - P_{\text{obs}, i} \right)^2
\]

The optimization yields the following fitted parameters:

\[
P_0^\ast = \text{[fitted value]}, \quad \bar{P}^\ast = \text{[fitted value]}, \quad \lambda^\ast = \text{[fitted value]}
\]

These fitted parameters provide a close match between the observed data and the model and helps us understand for better decision-making in resource allocation.

\section{Experimental Details}

\subsection{Finetuning Settings}

% [An rebuttal 1-sentence for why using LoRA as opposed to full finetuning] %Since the study focus on mainly the question of FLOPs efficiency instead of. 

All experiments were conducted with parameter-efficient finetuning method LoRA \cite{hu2021lora}. For the LoRA adapter, we specified a rank of 128, an $\alpha$ value of 512, and a dropout rate of 0.1 and applied it across all attention matrices. Adding the LoRA adapter introduce minimal FLOPs overhead during training---having no impact on our FLOPS analysis---and mainly reduce memory requirements for more accessible training.

We follow standard practices in LLM finetuning \citet{wang2023far, ivison2023camels} and use the AdamW optimizer with beta-parameters $(\beta_1, \beta_2) = (0.9, 0.99)$. 

% [explanation for using larger learning rate in small percentage 1-5\%, and smaller learning rate in large percentage, the goal is to produce the optimal train loss and is controlled.]

The learning rate is set to 2e-5 for the 7B/8B/13B models and 1e-5 for the 70B models. For data budget \{2.5\%,5\%\}, we double the learning rate to ensure convergence in loss. For all experiments, we use a warmup ratio of 0.03, BFloat16 precision, and an effective batch size of 128. For 70B model training, we used QLoRA to reduce the memory requirements and speedup the training. 

For smaller data budget (2.5\%-5\%) experiments, we perform five trials across five random seeds. For larger data budget (10\%-100\%) experiments, we perform three trials across three random seeds. We report mean target task performance in our analysis. Optimization seeds are controlled through the entirety of the experiments.

\subsection{Pretrained Models}
% \begin{wraptable}{r}{1.0\textwidth}
    % \vspace{-25pt}
\begin{table}[b]
    \centering
    \centering
    \begin{tabular}{lcc}
        \toprule
        \textbf{Base LMs} & \textbf{\# Params} & \textbf{\# Tokens} \\
        \midrule
        \multirow{3}{*}{\centering \llama} & 6.7B & 2.0T \\
        & 13.0B & 2.0T \\
        & 65.2B & 2.0T \\
        \midrule
        \textsc{Llama-3} & 8B & 15.6T \\
        \bottomrule
    \end{tabular}
    \caption{Base models that we finetuned in this work.}
\label{tab:base_models}
\end{table}

% \end{wraptable}

\Cref{tab:base_models} lists the pre-trained models we finetuned in this work. We expect our findings to generalize to these models and future, stronger open base models.

\subsection{Training Datasets}
For training, we use the same four processed datasets as in \cite{wang2023far, ivison2023camels}, all of which are either human-authored or annotated. Details are provided in \Cref{tab:dataset_details}. The FLAN V2 and COT datasets are derived from existing NLP datasets, while DOLLY and OPEN ASSISTANT 1 contain open-ended generation examples with human-written responses. These datasets vary in format, sequence length, and tasks. Following \cite{wang2023far}, we standardize their format using the ‘Tulu’ structure.

\subsection{Evaluation Datasets}

We evaluate our method on three benchmark datasets: MMLU \citep{hendrycks2020measuring}, BBH \citep{suzgun2022challenging}, and  IFEval \citep{zhou2023instruction}. \Cref{tab:evaluation_data} contains more details about each tasks. Each subtask comes with few-shot examples or sample responses, which are used as validation set $\mathcal{V}$ for data selection and as few-shot in-context learning demonstration in evaluation.  

An important aspect of the data selection approach is the size of the validation set used for computing utility scores. While a small validation set might cause overfitting or insufficient representation of the target task, our experiments indicate that even modestly sized validation sets (e.g., containing more than 50 examples) are sufficient for methods like BM25 and Embed to perform effectively. Since we are selecting large subsets from the training data, the precision required from the validation set is not overly stringent. These methods can effectively capture relevant training samples without significant risk of overfitting, suggesting that, in practice, reasonably sized validation sets suffice for similarity-based data selection in LLM finetuning.

\begin{table}[t]
\centering
\resizebox{\textwidth}{!}{
\begin{tabular}{lclcccc}
\toprule
\textbf{Dataset} & \textbf{\# Instance} & \textbf{Sourced from} & \textbf{\# Rounds} & \textbf{Prompt Len.} & \textbf{Completion Len.} \\ \midrule
\textsc{Flan V2} & 100,000 & NLP datasets and human-written instructions & 1 & 355.7 & 31.2 \\ 
\textsc{CoT} & 100,000 & NLP datasets and human-written CoTs & 1 & 266 & 53.2 \\ 
\textsc{Dolly} & 15,011 & Human-written from scratch & 1 & 118.1 & 91.3 \\ 
\textsc{Open Assistant 1} & 55,668 & Human-written from scratch & 1.6 & 34.8 & 212.5 \\ 
\midrule
\multicolumn{2}{c}{\textsc{Total Number of Tokens: 95.7 Million}}  \\
\end{tabular}
}
\caption{Details of training dataset from \cite{wang2023far}. Len. is short for token length.}
\label{tab:dataset_details}
\end{table}

\begin{table}[!t]
\centering
\begin{tabular}{lccccc}
\hline
\textbf{Dataset} & \textbf{\# Shot} & \textbf{\# Tasks} & $\vert \boldsymbol{\mathcal{V}} \vert$ & $\vert \boldsymbol{\mathcal{T}} \vert$ & \textbf{Answer Type} \\ \hline
MMLU & 5 & 57 & 285 & 18,721 & Letter options \\
BBH & 3 & 23 & 69 & 920 & COT and answer \\
IFEval & 1 & - & 50 & 541 & Open Generation \\ \hline
\end{tabular}
\caption{Statistics of evaluation datasets. The selection of evaluation tasks cover different kinds of answer types.}
\label{tab:evaluation_data}
\end{table}

\section{Results: LLAMA3}

\begin{figure}[!t]
    \centering
    \resizebox{1.0\textwidth}{!}{%
    \includegraphics{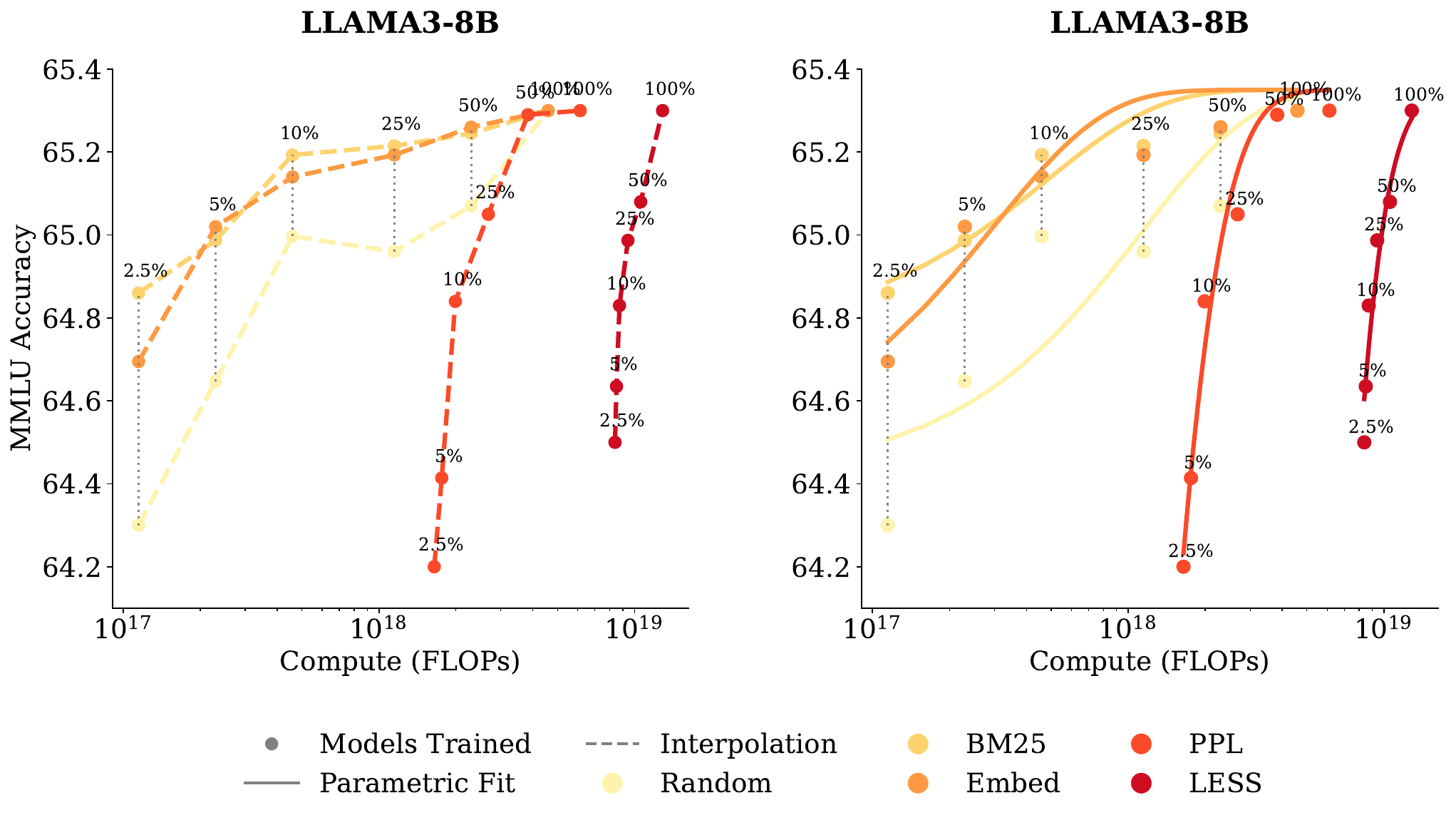} 
    }
    \caption{\textbf{Performance and Parametric Fit of Performance With Compute-Constrained Data Selection.} \textit{(Left)} We show all of our different runs
for a given model size, where each scatter point is the final target task performance of a single run. \textit{(Right)} We fit a parametric model of the performance in \Cref{eq:parametric-function} and display that as curves to pair with the empirical results as scatter points. 
    }
    \label{fig:llama3-results}
\end{figure}

We plot additional results on target task MMLU using \textsc{Llama}3 8B model in \Cref{fig:llama3-results}. Similar to \textsc{Llama}2, \textsc{Llama}3 8B results show that cheaper lexicon-based (BM25) and embedding-based (Embed) methods significantly outperform perplexity-based (PPL) and gradient-based (LESS) method. The marginal gains from using more sophisticated methods do not justify their selection costs.

\label{appendix-results-llama3}

\begin{figure}[!t]
    \centering
    \resizebox{1.0\textwidth}{!}{%
    \includegraphics{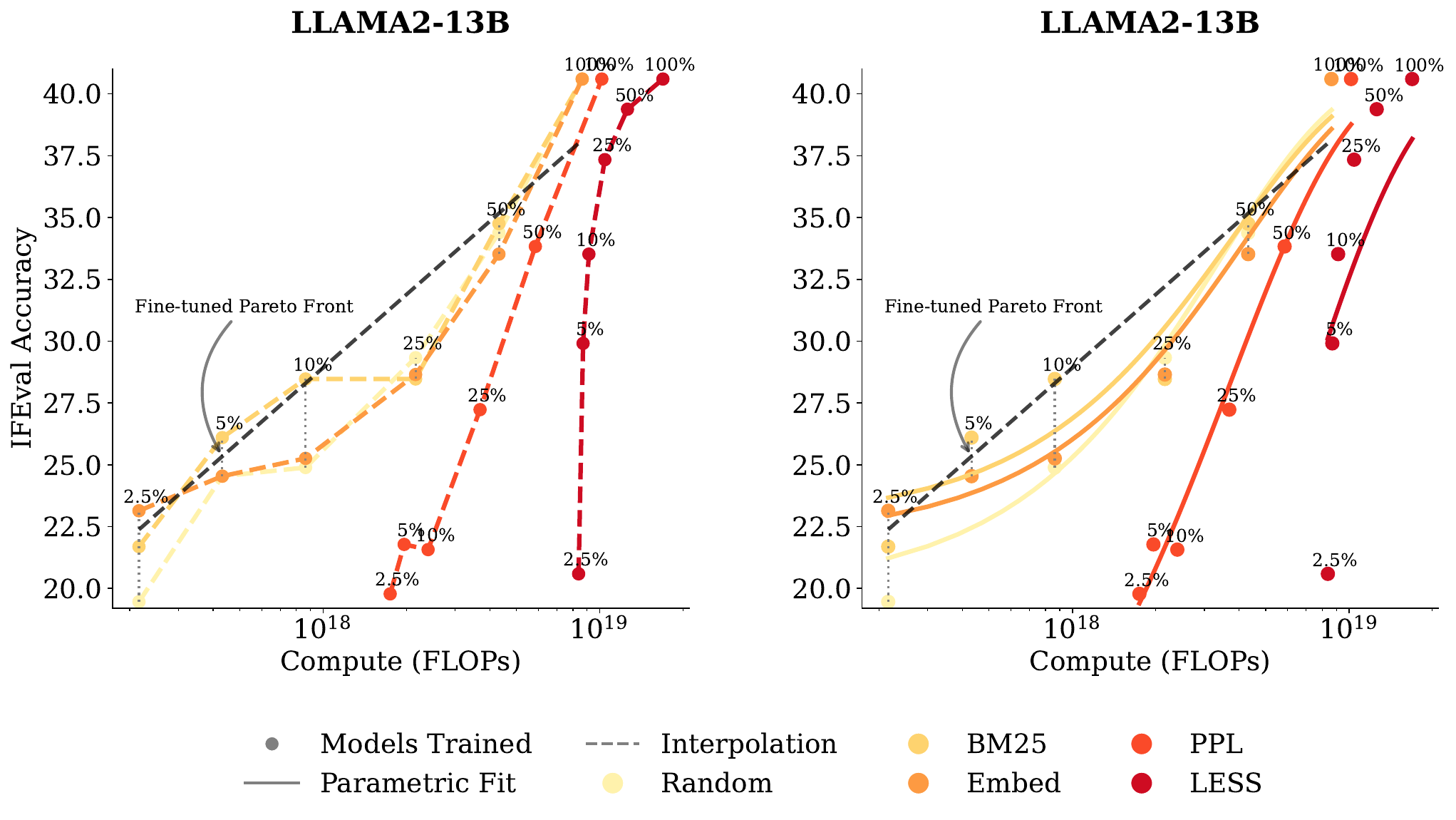} 
    }
    \caption{\textbf{Performance and Parametric Fit of Performance With Compute-Constrained Data Selection} on IFEval. \textit{(Left)} We show all of our different runs
for a given model size, where each scatter point is the final target task performance of a single run. \textit{(Right)} We fit a parametric model of the performance in \Cref{eq:parametric-function} and display that as curves to pair with the empirical results as scatter points. We fit a pareto front in dashed line based on these optimal strategies. At medium compute budgets, cheaper data selection methods outperform PPL and LESS, which rely on model information.
    }
    \label{fig:llama2-13b-ifeval-results}
\end{figure}

% While Random

% LESS might not be transferable between different model families. 
\section{Additional Results: IFEval}
\label{Appendix-Additional-Results-IFEval}
We plot additional results on target task IFEval with \textsc{Llama}3 13B model in \Cref{fig:llama2-13b-ifeval-results}. As models scale to 13B, expensive data selection methods still underperform, despite their relative cost diminishing with larger models. Cheaper methods remain preferred.

\section{Extrapolating from the Parametric Function}
\label{Appendix-Extrapolation}

\begin{figure}[!t]
    \centering
    \resizebox{1.0\textwidth}{!}{%
    \includegraphics{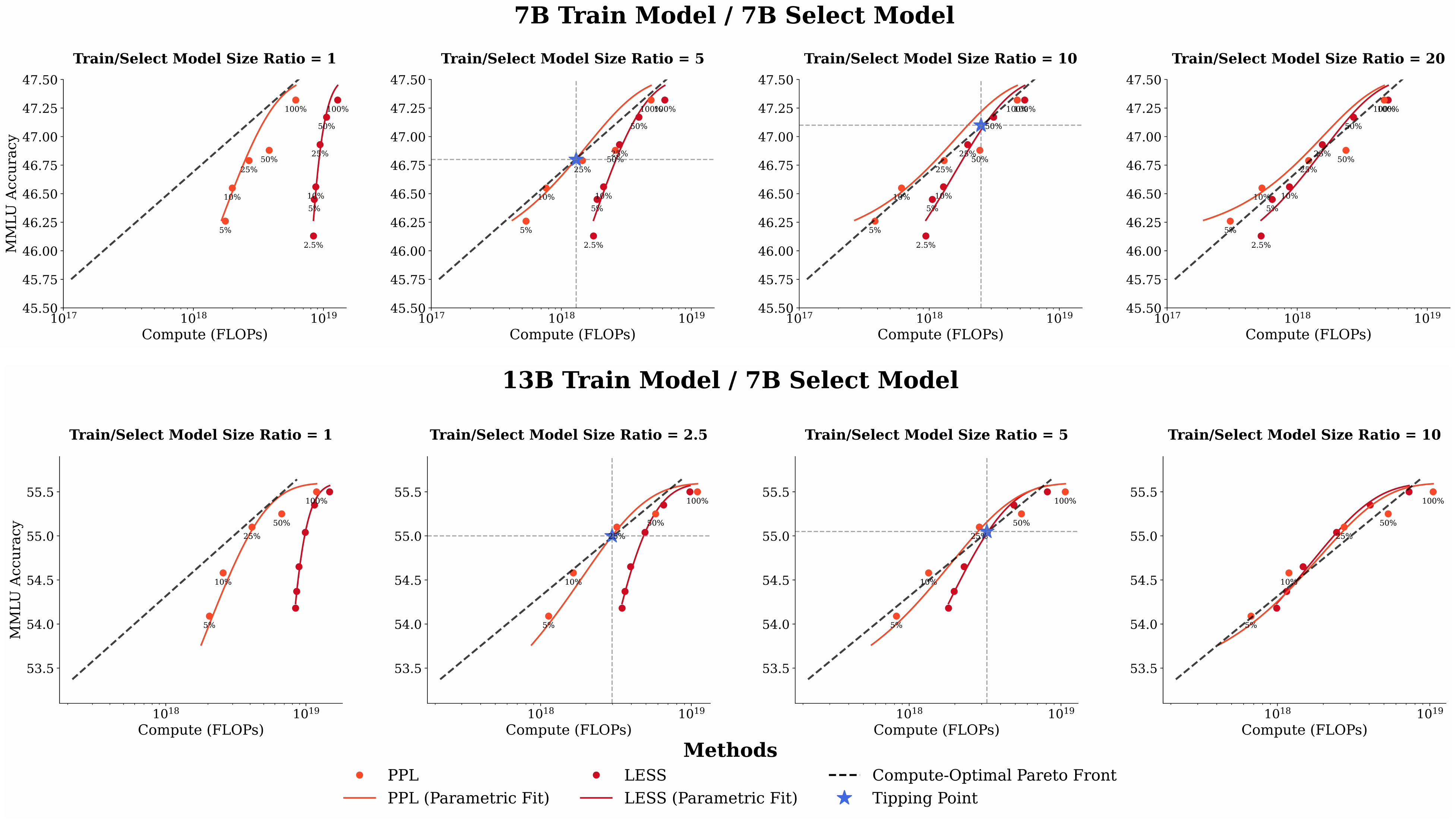} 
    }
    \caption{\textbf{Extrapolating Compute-Optimal Model Size for MMLU}. By extrapolating from the parametric fits obtained from the 7B and 13B model results for MMLU, we can find the compute-optimal ratio between the training model size and the selector model size required for the perplexity-based and gradient-based method. For perplexity-based data selection, our extrapolation suggests that training model should be larger (5x) than the selector model—around 35B model parameters. For gradient-based data selection to be compute-optimal, our extrapolation suggest that the training model should be significantly larger (10x) than the selector model—specifically, around 70B parameters in this case.}
    \label{fig:extrapolation-mmlu}
\end{figure}

\begin{figure}[!t]
    \centering
    \resizebox{1.0\textwidth}{!}{%
    \includegraphics{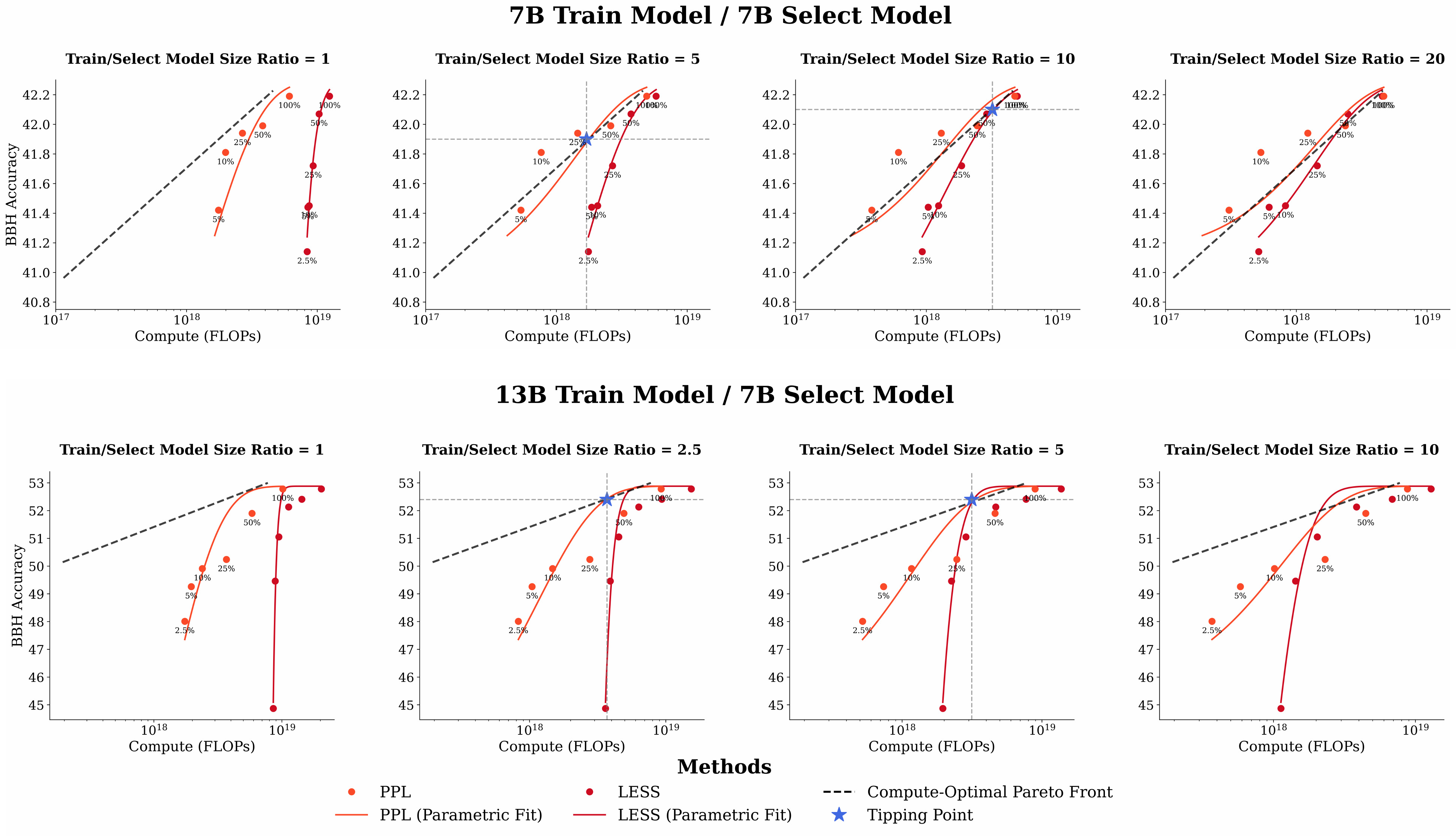} 
    }
    \caption{\textbf{Extrapolating Compute-Optimal Model Size for BBH}.  We perform a similar extrapolation for BBH and find results consistent with those in \Cref{fig:extrapolation-mmlu}. Specifically, for the perplexity-based method to be compute-optimal, the training model should be 5x larger than the selection model, and for the gradient-based method, the training model should be approximately 10x larger than the selection model.}
    \label{fig:extrapolation-bbh}
\end{figure}

In \Cref{fig:extrapolation-mmlu} and \Cref{fig:extrapolation-bbh}, we extrapolate the parametric fits obtained from our 7B and 13B model runs by increasing the ratio between the training model size and the data selection model size, thereby effectively reducing the proportional cost of gradient-based data selection. Graphically, this is seen when the parametric fit crosses the existing compute-optimal frontier—a tipping point at which the method becomes compute-optimal.

We found that the training-to-selection model size ratio should be approximately 5 for perplexity-based data selection, and 10 for gradient-based data selection. This suggests that using more expensive data selection methods becomes advantageous when the training model size largely exceeds the selection model size by the compute-optimal ratio. Under these conditions, meaningful efficiency gains can be achieved compared to cheaper data selection methods.

\section{Additional Analysis: Multiple Task-Specific Models.}
\label{Appendix-Break-Even-Analysis}

\begin{figure}[!t]
    \centering
    \resizebox{1.0\textwidth}{!}{%
    \includegraphics{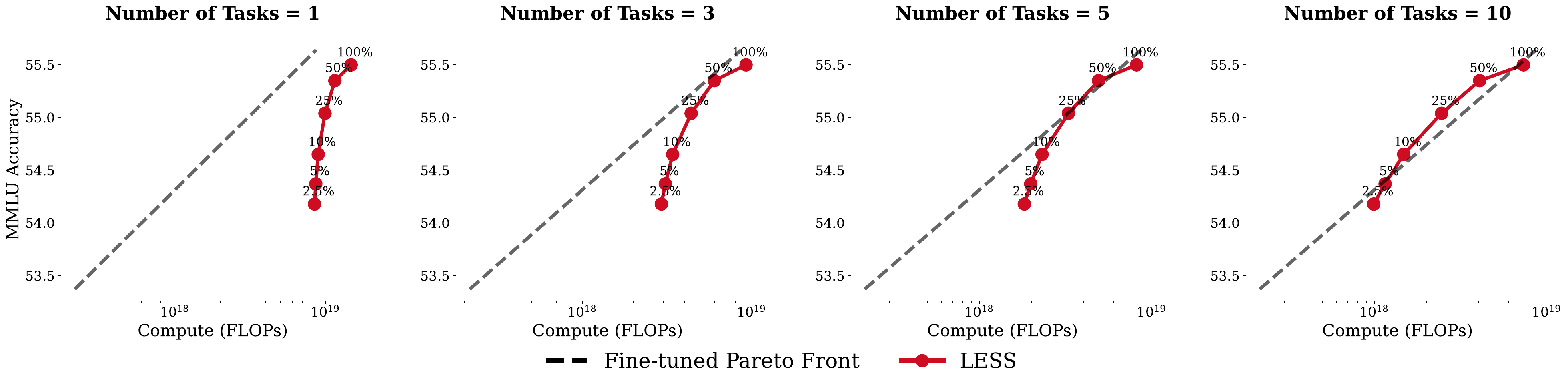} 
    }
    \caption{\textbf{Multiple Task-Specific Model Break-Even Analysis} for 13B MMLU. Performance under compute-constraints reach the finetuned Pareto frontier at 5 tasks, surpassing it at 10 tasks.}
    \label{fig:break-even-analysis-2}
\end{figure}

\begin{figure}[!t]
    \centering
    \resizebox{1.0\textwidth}{!}{%
    \includegraphics{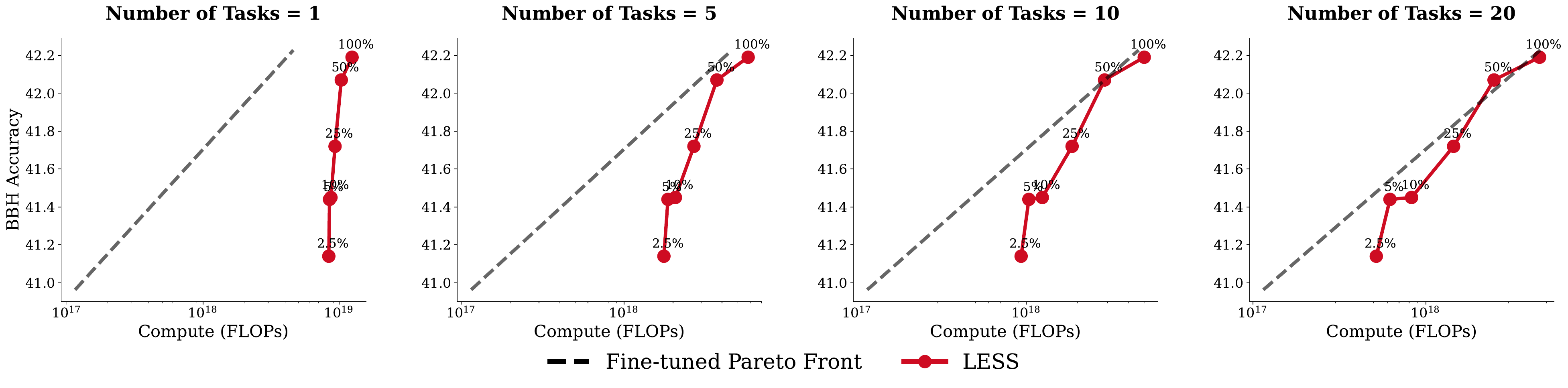} 
    }
    \caption{\textbf{Multiple Task-Specific Model Break-Even Analysis} for 7B BBH. Performance under compute-constraints reach the finetuned Pareto frontier at 10 tasks.}
    \label{fig:break-even-analysis-3}
\end{figure}

\begin{figure}[!t]
    \centering
    \resizebox{1.0\textwidth}{!}{%
    \includegraphics{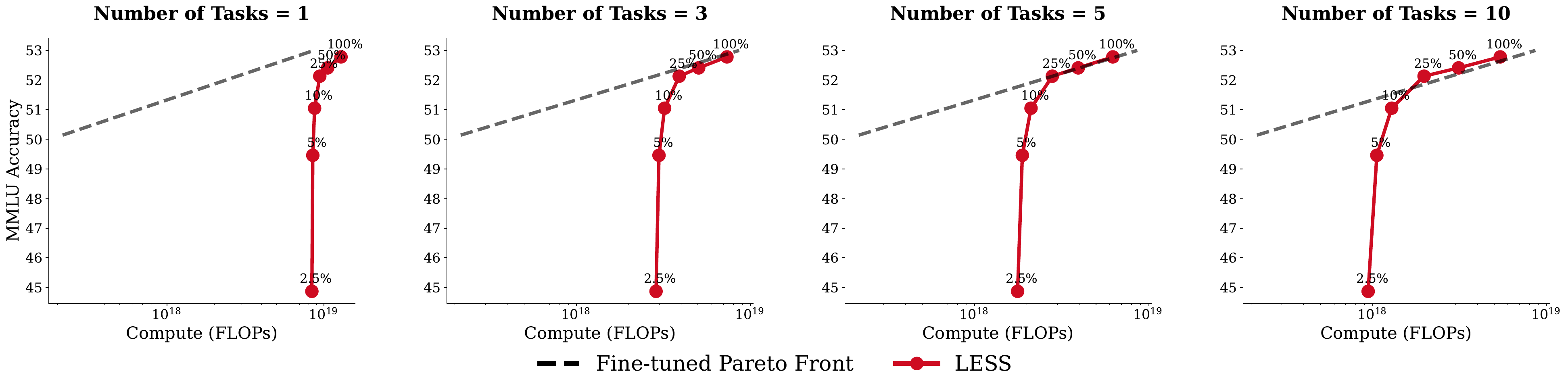} 
    }
    \caption{\textbf{Multiple Task-Specific Model Break-Even Analysis} for 13B BBH. Performance under compute-constraints reach the finetuned Pareto frontier at 5 tasks, surpassing it at 10 tasks.}
    \label{fig:break-even-analysis-4}
\end{figure}

\begin{figure}[!t]
    \centering
    \resizebox{1.0\textwidth}{!}{%
    \includegraphics{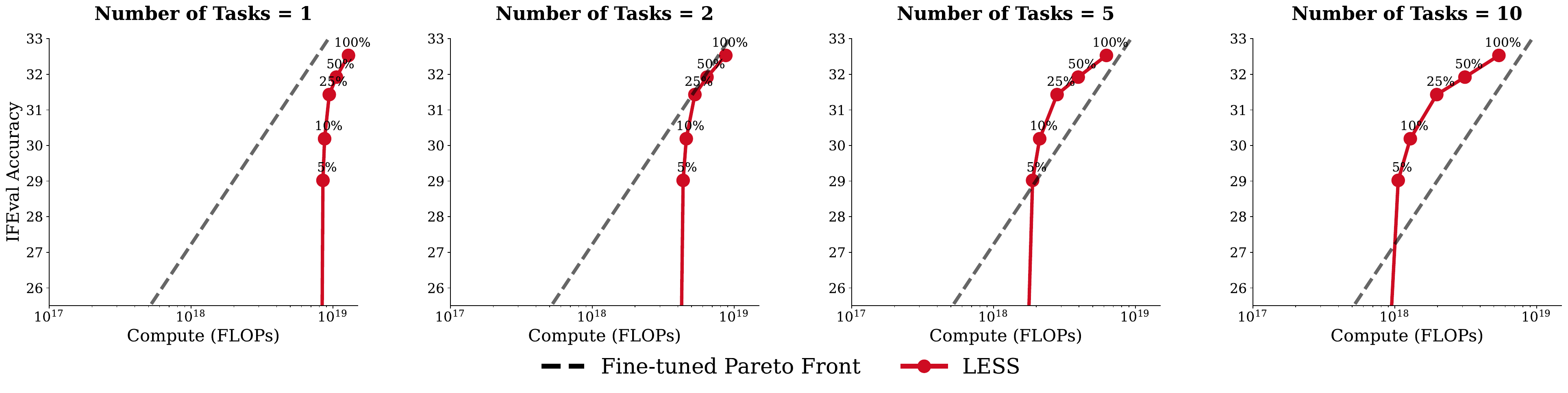} 
    }
    \caption{\textbf{Multiple Task-Specific Model Break-Even Analysis} for 7B IFEval. Performance under compute-constraints reach the finetuned Pareto frontier at 5 tasks, surpassing it at 10 tasks.}
    \label{fig:break-even-analysis-5}
\end{figure}

\begin{figure}[!t]
    \centering
    \resizebox{1.0\textwidth}{!}{%
    \includegraphics{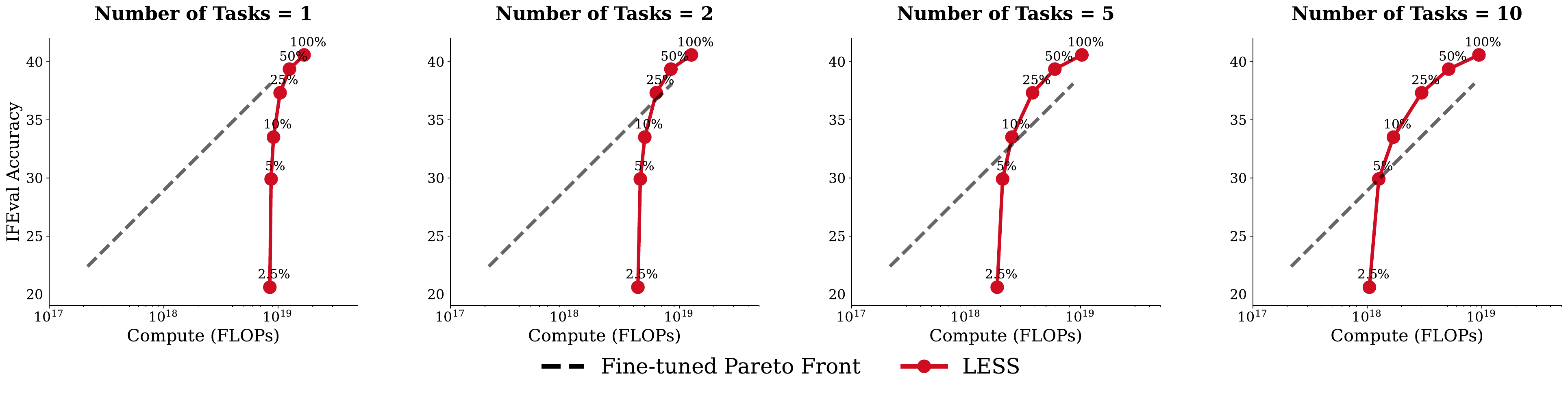} 
    }
    \caption{\textbf{Multiple Task-Specific Model Break-Even Analysis} for 13B IFEval. Performance under compute-constraints reach the finetuned Pareto frontier at 2 tasks, surpassing it at 5, 10 tasks.}
    \label{fig:break-even-analysis-6}
\end{figure}

We perform break-even analysis by varying the number of target tasks for each method. \Cref{fig:break-even-analysis-2} presents the analysis for the 13B model on MMLU, while \Cref{fig:break-even-analysis-3} focuses on the 7B model on BBH and \Cref{fig:break-even-analysis-4} examines the 13B model on BBH. Additionally, \Cref{fig:break-even-analysis-5} and \Cref{fig:break-even-analysis-6} provide results for the 7B and 13B models on IFEval, respectively.

% Show for 13B models.

\section{Additional Analysis: Data Similarity}
\label{Appendix-Data-Similarity-Analysis}

\begin{figure}[!t]
    \centering
    \resizebox{0.85\textwidth}{!}{%
    \includegraphics{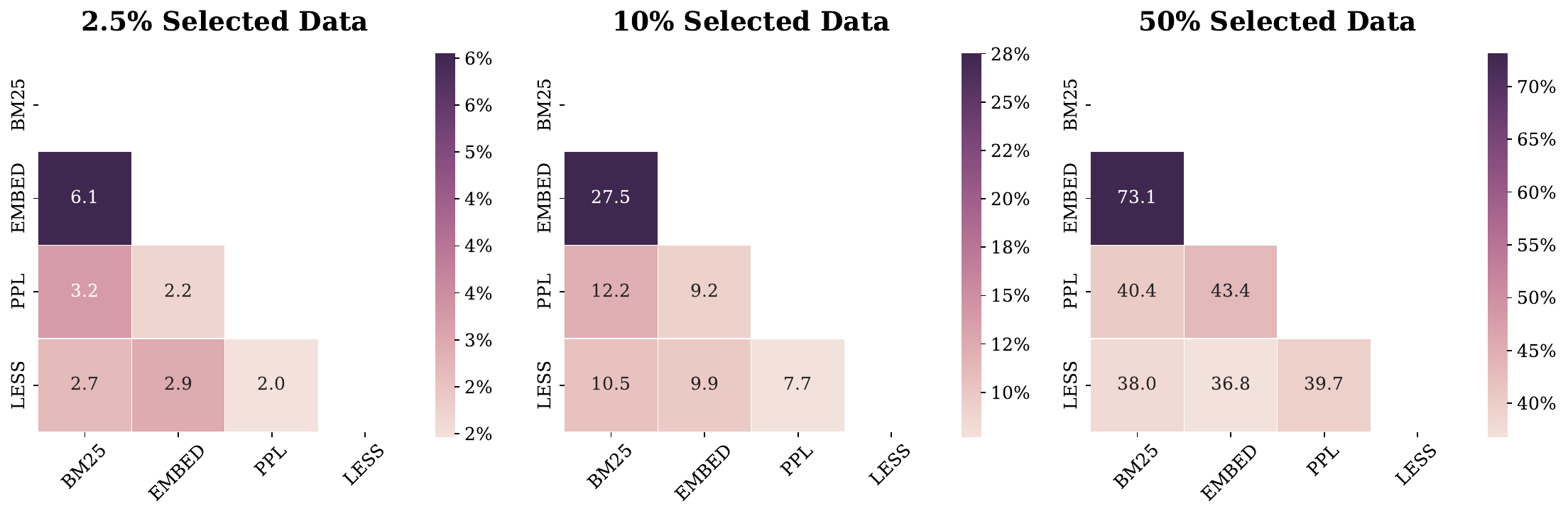} 
    }
    \caption{\textbf{Data Similarity Between Data Selection Methods} for BBH. }
    \label{fig:data-similarity-bbh}
\end{figure}

\begin{figure}[!t]
    \centering
    \resizebox{0.85\textwidth}{!}{%
    \includegraphics{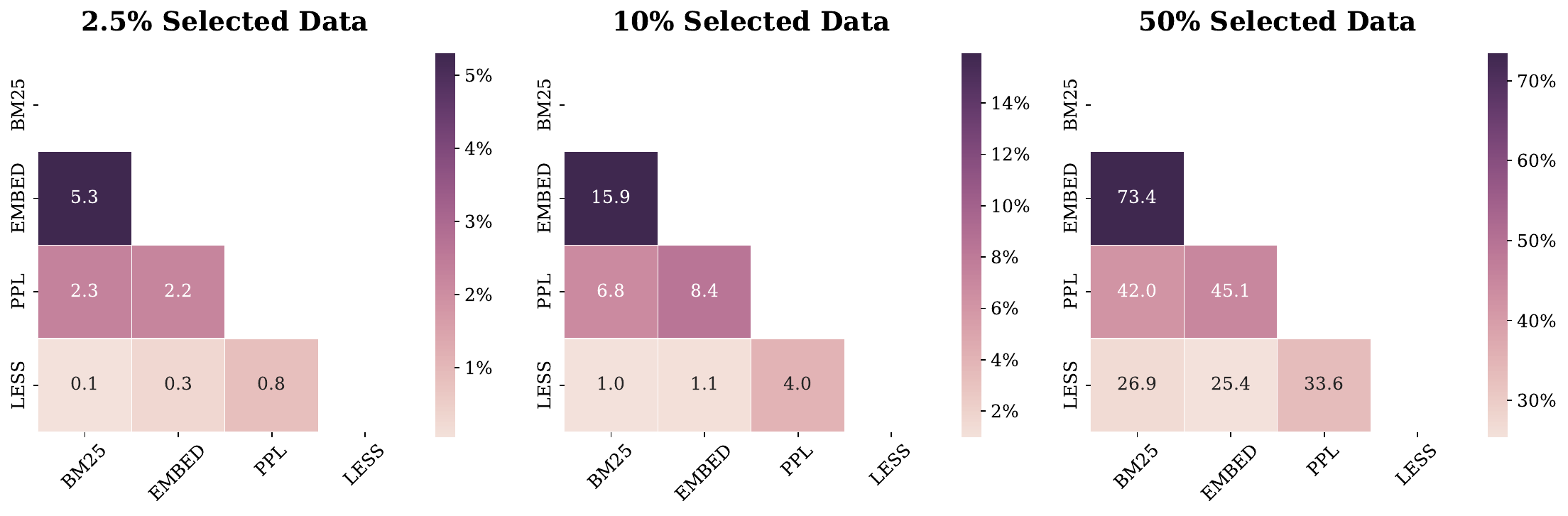} 
    }
    \caption{\textbf{Data Similarity Between Data Selection Methods} for IFEval. }
    \label{fig:data-similarity-ifeval}
\end{figure}

\Cref{fig:data-similarity-bbh} and \Cref{fig:data-similarity-ifeval} show the Jaccard similarity of data selected by different data selection methods for the BBH and IFEval target tasks, respectively. Across various percentages of selected data, Embedding and BM25 exhibit the highest similarity. In contrast, LESS shares the least similarity with the other data selection methods.

\section{Limitations}

\paragraph{Repeating Finetuning Data} In this work, we focus on finetuning the entire dataset for only one epoch. With a larger compute budget, it is possible to repeat fractions or entire datasets multiple times. Multi-epoch settings, as explored in \citep{xia2024less}, could potentially provide further training speedup by repeating data selectively.

\paragraph{Sensitivity to Hyperparameters} The effectiveness of finetuning can be highly sensitive to hyperparameters such as learning rate, dropout, or optimizer choice. There may be a specific learning rate that leads to quicker convergence. In this work, we fixed most hyperparameters to commonly used values for fine-tuning LLMs, leaving further exploration of hyperparameter tuning for future work.

\paragraph{Other Data Selection Methods} There are additional data selection methods not covered in this work that warrant investigation. While we focused on methods in terms of their compute efficiency, other approaches, such as classifier-based methods, could offer insights and deserve further exploration.

\ificlrfinal
% No need to include as we will rid of the anonymity 
\else
\section{Reproducibility Statement}
To ensure reproducibility, we will release all code and notebooks used for training and plotting. This includes instructions for replicating our experiments and generating the visualizations presented in this paper. Our aim is to make it straightforward for others to reproduce and build upon our findings.
\fi

\end{document}